\documentclass[10pt,journal,compsoc]{IEEEtran}

\ifCLASSOPTIONcompsoc
  \usepackage[nocompress]{cite}
\else
  \usepackage{cite}
\fi

\usepackage{xcolor}
\usepackage{colortbl}
\usepackage{xspace}
\usepackage{xcolor,colortbl}
\usepackage{graphicx}
\usepackage{amsmath}
\usepackage{amssymb}
\usepackage{enumerate}
\usepackage[
  pagebackref=true,
  breaklinks=true,
  letterpaper=true,
  colorlinks,
  bookmarks=false]{hyperref}
\usepackage{tikz}

\pdfoutput=1

\definecolor{attr}{rgb}{0.85,0.88,0.90}
\definecolor{code}{rgb}{0.95,0.93,0.90}
\definecolor{Red}{rgb}{1,0.0}
\definecolor{green}{rgb}{0,1,0}
\definecolor{lightred}{rgb}{1, 0.4, 0.4}
\definecolor{lightblue}{rgb}{0.6, 07, 0.9}
\definecolor{lightyellow}{rgb}{1, 1, 0.6}

\definecolor{Gray}{gray}{0.5}
\definecolor{LightCyan}{rgb}{0.88,0.1,0.1}
\newcolumntype{a}{>{\columncolor{Gray}}c}
\newcolumntype{b}{>{\columncolor{white}}c}
  
\newcommand{\off}[1]{}

\newcommand{\fsift}{f_{\operatorname{sift}}}
\newcommand{\real}{\mathbb{R}}

\newcommand{\bcnn}{B-CNN\xspace}

\newcommand{\netvlad}{NetVLAD\xspace}
\newcommand{\netbovw}{NetBoVW\xspace}
\newcommand{\netfv}{NetFV\xspace}

\newcommand*{\ie}{\textit{i.e.}\@\xspace}
\newcommand*{\eg}{\textit{e.g.}\@\xspace}
\newcommand*{\etc}{\textit{etc.}\@\xspace}
\newcommand{\etal}{\textit{et al}.}
\newcommand{\na}{\textit{n/a}}

\graphicspath{{./fig/patch_strip/}{./fig/confusing_imgs/}{./fig/bcnn-patches/}}

\tikzset{
 image label/.style={
   fill=white,
   text=black,
   font=\footnotesize,
   anchor=south east,
   xshift=-0.02cm,
   yshift=0.02cm,
   at={(0,0)}
 }
}

\newcommand{\puti}[2]
{%
\begin{tikzpicture}
\node[anchor=south east,inner sep=0] at (0,0) {#2};
\node[image label]{#1};
\end{tikzpicture}%
}
\begin{document}
\title{Bilinear CNNs for Fine-grained Visual Recognition}

\author{Tsung-Yu~Lin~~~~~
        Aruni~RoyChowdhury~~~~~
        Subhransu~Maji% <-this % stops a space
\IEEEcompsocitemizethanks{\IEEEcompsocthanksitem T.-Y. Lin, A. RoyChowdhury, and S. Maji are with the College of Information and Computer Sciences, University of Massachusetts Amherst, USA. E-mails: \{tsungyulin, arunirc, smaji\}@cs.umass.edu}
}
\IEEEtitleabstractindextext{
\begin{abstract}
We present a simple and effective architecture for fine-grained visual recognition called \emph{Bilinear Convolutional Neural Networks (B-CNNs)}.
These networks represent an image as a \emph{pooled outer product} of features derived from two CNNs and capture localized feature interactions in a translationally invariant manner.
B-CNNs belong to the class of orderless texture representations but unlike prior work they can be trained in an end-to-end manner.
Our most accurate model obtains 84.1\%, 79.4\%, 86.9\% and 91.3\% per-image accuracy on the Caltech-UCSD birds~\cite{WahCUB_200_2011}, NABirds~\cite{van2015building}, FGVC aircraft~\cite{maji2013fine}, and Stanford cars~\cite{krause20133d} dataset respectively and runs at 30 frames-per-second on a NVIDIA Titan X GPU.
We then present a systematic analysis of these networks and show that (1) the bilinear features are highly redundant and can be reduced by an order of magnitude in size without significant loss in accuracy, (2) are also effective for other image classification tasks such as texture and scene recognition, and (3) can be trained from scratch on the ImageNet dataset offering consistent improvements over the baseline architecture.
Finally, we present visualizations of these models on various datasets using top activations of neural units and gradient-based inversion techniques.
The source code for the complete system is available at \url{http://vis-www.cs.umass.edu/bcnn}.
\end{abstract}
}

\maketitle
\IEEEdisplaynontitleabstractindextext
\IEEEpeerreviewmaketitle

\IEEEraisesectionheading{\section{Introduction}\label{sec:introduction}}

\IEEEPARstart{F}{ine-grained} recognition involves classification of instances within a subordinate category. Examples include recognition of species of birds, models of cars, or breeds of dogs. 
These tasks often require recognition of highly localized attributes of objects while being invariant to their pose and location in the image.
For example, distinguishing a ``California gull" from a ``Ringed-bill gull" requires the recognition of patterns on their bill, or subtle color differences of their feathers~\cite{gull-diff}. There are two  broad classes of techniques that are effective for these tasks. 
\emph{Part-based models} construct representations by localizing parts and extracting features conditioned on their detected locations. This makes subsequent reasoning about appearance easier since the variations due to location, pose, and viewpoint changes are factored out. \emph{Holistic models} on the other hand construct a representation of the entire image directly. These include classical image representations, such as Bag-of-Visual-Words~\cite{csurka04visual} and their variants popularized for texture analysis.
Most modern approaches are based on representations extracted using Convolutional Neural Networks (CNNs) pre-trained on the ImageNet dataset~\cite{russakovsky15imagenet}. While part-based models based on CNNs are more accurate, they require part annotations during training. This makes them less applicable in domains where such annotations are difficult or expensive to obtain, including categories without a clearly defined set of parts such as textures and scenes.
 
In this paper we argue that the effectiveness of part-based reasoning is due to their invariance to position and pose of the object. 
Texture representations are translationally invariant by design as they are based on aggregation of local image features in an \emph{orderless} manner. While classical texture representations based on SIFT~\cite{lowe99object} and their recent extensions based on CNNs~\cite{cimpoi2016},\cite{gong14multi-scale}, have been shown to be effective at fine-grained recognition, they have not matched the performance of part-based approaches.
A potential reason for this gap is that the underlying features in texture representations are not learned in an end-to-end manner and are likely to be suboptimal for the recognition task. 

We present \emph{Bilinear CNNs (B-CNNs)} that address several drawbacks of existing deep texture representations. 
Our key insight is that several widely-used texture representations can be written as a pooled outer product of two suitably designed features.
When these features are based on CNNs the resulting architecture consists of standard CNN units for feature extraction, followed by a specially designed \emph{bilinear layer} and a \emph{pooling layer}.
The output is a fixed high-dimensional representation which can be combined with a fully-connected layer to predict class labels. 
The simplest bilinear layer is one where two \emph{identical} features are combined with an outer product. 
This is closely related to the \emph{Second-Order Pooling} approach of Carreira \etal~\cite{carreira2012semantic} popularized for semantic image segmentation. 
We also show that other texture representations can be written as B-CNNs once suitable non-linearities are applied to the underlying features.
This results in a family of layers which can be plugged into existing CNNs for end-to-end training on large datasets, or domain-specific fine-tuning for transfer learning.
B-CNNs outperform existing models, including those trained with part-level supervision, on a variety of fine-grained recognition datasets. Moreover, these models are fairly efficient. Our most accurate model implemented in MatConvNet~\cite{vedaldi15matconvnet} runs at 30 frames-per-second on a NVIDIA Titan X GPU and obtains \textbf{84.1\%}, \textbf{79.4\%}, \textbf{86.9\%} and \textbf{91.3\%} per-image accuracy on Caltech-UCSD birds~\cite{WahCUB_200_2011}, NABirds~\cite{van2015building}, FGVC aircraft~\cite{maji2013fine}, and Stanford cars~\cite{krause20133d} dataset respectively.
 
This manuscript combines the analysis of our earlier works~\cite{lin2015bilinear,Lin2016CVPR} and extends them in a number of ways. We present an account of related work, including extensions published subsequently (Section~\ref{s:related}).
We describe the B-CNN architecture (Section~\ref{s:model}), and 
present a unified analysis of exact and approximate end-to-end trainable formulations of Second-Order Pooling (O2P), Fisher Vector (FV), Vector-of-Locally-Aggregated Descriptors (VLAD), Bag-of-Visual-Words (BoVW) in terms of their accuracy on a variety of fine-grained recognition datasets (Section~\ref{s:texture}-\ref{s:experiments}).
We show that the approach is general-purpose and is effective at other image classification tasks such as material, texture, and scene recognition (Section~\ref{s:experiments}).
We present a detailed analysis of dimensionality reduction techniques and provide trade-off curves between accuracy and dimensionality for different models, including a direct comparison with the recently proposed \emph{compact bilinear pooling} technique~\cite{Gao2016CVPR} (Section~\ref{sec:dim_red}). 
Moreover, unlike prior texture representations based on networks pre-trained on the ImageNet dataset, B-CNNs can be trained from scratch and offer consistent improvements over the baseline architecture with a modest increase in the computation cost (Section~\ref{s:imagenet}).
Finally we visualize the top activations of several units in the learned models and apply the gradient-based technique of Mahendran and Vedaldi~\cite{mahendran16visualizing} to visualize inverse images on various texture and scene datasets (Section~\ref{s:viz}).
We have released the complete source code for the system at \url{http://vis-www.cs.umass.edu/bcnn}.
\section{Related work}\label{s:related}
\textbf{Fine-grained recognition techniques.} After AlexNet's~\cite{krizhevsky12imagenet} impressive performance on the ImageNet classification challenge, several authors (\eg, ~\cite{donahue13decaf,razavin14cnn-features}) have demonstrated that features extracted from layers of a CNN are effective at fine-grained recognition tasks. Building on prior work on part-based techniques (\eg,~\cite{bourdev2011describing, farrell2011birdlets, zhang2012pose}), Zhang \etal~\cite{zhang14part-based}, and Branson \etal~\cite{branson14bird} demonstrated the benefits of combining CNN-based part detectors~\cite{girshick14rich} and CNN-based features for fine-grained recognition tasks. Other approaches use segmentation to guide part discovery in a weakly-supervised manner and train part-based models~\cite{krause2015fine}. Among the non part-based techniques, texture descriptors such as FV and VLAD have traditionally been effective for fine-grained recognition. For example, the top performing method on FGCOMP'12 challenge used SIFT-based FV representation~\cite{gosselin2014revisiting}.

Recent improvements in deep architectures have also resulted in improvements in fine-grained recognition. These include architectures that have increased depth such as the ``deep"~\cite{chatfield14return} and ``very deep"~\cite{simonyan14very} networks from the Oxford's VGG group, inception networks~\cite{Szegedy_2015_CVPR}, and ``ultra deep" residual networks~\cite{He_2016_CVPR}. Spatial Transformer Networks~\cite{jaderberg15spatial} augment CNNs with parameterized image transformations and are highly effective at fine-grained recognition tasks. Other techniques augment CNNs with ``attention" mechanisms that allow focused reasoning on regions of an image~\cite{recurrentAttention, ba2015multiple}. B-CNNs can be viewed as an implicit spatial attention model since the outer product modulates one feature based on the other, similar to the multiplicative feature interactions in attention mechanisms. Although not directly comparable, Krause \etal~\cite{krause2016unreasonable} showed that the accuracy of deep networks can be improved significantly by using two orders of magnitude more training data obtained by querying category labels on search engines. Recently, Moghimi~\etal~\cite{MoghimiBMVC2016} showed boosting B-CNNs offers consistent improvements on fine-grained tasks.

\textbf{Texture representations and second-order features.} Texture representations have been widely studied for decades. Early work~\cite{leung2001representing} represents the texture by computing the statistics of linear filter-bank responses (\eg, wavelets and steerable pyramids). The use of second-order features of filter-bank responses was pioneered by Portilla and Simoncelli~\cite{portilla2000parametric}. Recent variants such as FV~\cite{perronnin07fisher} and O2P~\cite{carreira2012semantic} with SIFT were shown to be a highly effective for image classification and semantic segmentation~\cite{everingham07pascal} tasks respectively.

The advantages of combining orderless texture representations and deep features have been studied in a number of recent works.
Gong~\etal~performed a \emph{multi-scale orderless pooling} of CNN features~\cite{gong14multi-scale} for scene classification. 
Cimpoi \etal~\cite{cimpoi2016} performed a systematic analysis of texture representations by replacing linear filter-banks with non-linear filter-banks derived from a CNN and showed it results in significant improvements on various texture, scene, and fine-grained recognition tasks. 
They found that orderless aggregation of CNN features was more effective than the commonly-used fully-connected layers on these tasks. 
However, a drawback of these approaches is that the filter banks are not trained in an end-to-end manner. 
Our work is also related to the \emph{cross-layer pooling} approach of Liu \etal~\cite{liu2016cross} who showed that second-order aggregation of features from two different layers of a CNN is effective at fine-grained recognition. 
Our work showed that feature normalization and domain-specific fine-tuning offers additional benefits, improving the accuracy from 77.0\% to 84.1\% using identical networks on the Caltech-UCSD Birds dataset~\cite{WahCUB_200_2011}.
Another subsequently published work of interest is the \emph{NetVLAD} architecture~\cite{Arandjelovic16} which provides a end-to-end trainable approximation of VLAD. The approach was applied to image-based geolocation problem. We include a comparison of NetVLAD to other texture representations in Section~\ref{s:experiments}.

\textbf{Texture synthesis and style transfer.} Concurrent to our work, Gatys \etal~showed that the Gram matrix of CNN features is an effective texture representation and by matching the Gram matrix of a target image one can create novel images with the same texture~\cite{gatys2015texture} and transfer styles~\cite{Gatys_2016_CVPR}. While the Gram matrix is identical to a pooled bilinear representation when the two features are the same, the emphasis of our work is \emph{recognition} and \emph{not} \emph{generation}. This distinction is important since Ustyuzhaninov \etal~\cite{ustyuzhaninov2016texture} show that the Gram matrix of a shallow CNN with random filters is sufficient for texture synthesis, while discriminative pre-training and subsequent fine-tuning are essential to achieve high performance for recognition.

\textbf{Polynomial kernels and sum-product networks.} An alternate strategy for combining features from two networks is to concatenate them and learn their pairwise interactions through a series of layers on top. However, doing this naively requires a large number of parameters since there are $O(n^2)$ interactions over $O(n)$ features requiring a layer with $O(n^3)$ parameters. Our explicit representation using an outer product has no parameters and is similar to a quadratic kernel expansion used in kernel support vector machines~\cite{scholkopf2001learning}. However, one might be able to achieve similar approximations using alternate architectures such as sum-product networks that efficiently model multiplicative interactions~\cite{gens2012discriminative}. 

\textbf{Bilinear model variants and extensions.} Bilinear models were used by Tanenbaum and Freeman~\cite{tenenbaum2000separating} to model two-factor variations such as ``style" and ``content" for images. While we also model two factor variations in location and appearance of parts, our goal is classification and not the explicit modeling of these factors. Our work is related to bilinear classifiers~\cite{pirsiavash2009bilinear} that express the classifier as a product of two low-rank matrices. Our models based on low dimensional representations described in Section~\ref{sec:dim_red} can be interpreted as bilinear classifiers. Our model is related to ``two-stream" architectures used to analyze videos where one network models the temporal aspect, while the other models the spatial aspect~\cite{simonyan14two-stream,fragkiadaki2015learning}. The idea of combining two features using the outer product has also been shown to be effective for other tasks such as visual question-answering~\cite{FukuiPYRDR16} where text and visual features are combined, action recognition~\cite{Feichtenhofer16} where optical flow and image features are combined.

\textbf{Low-dimensional bilinear features.} A drawback of the bilinear features is the memory overhead of storing the high-dimensional features. For example, the outer product of 512 dimensional features results in a 512$\times$512 dimensional representation. Our earlier work~\cite{lin2015bilinear} showed that the overall representation can be reduced to 512$\times$64 dimensions by projecting one of the features to a lower-dimensional space. Alternatively, the \emph{compact bilinear pooling}~\cite{Gao2016CVPR} applies \emph{tensor sketching}\cite{pham2013fast} to aggregate low-dimensional embeddings that approximate the bilinear features. In Section~\ref{sec:dim_red} we compare the two approaches and find that the projection method is simpler, faster, and equally effective. In most cases features size can be reduced 8-32$\times$ without significant loss in accuracy.

\textbf{Scalability and speed.} B-CNNs compare favorably to traditional CNN architectures in terms of speed since they replace several fully-connected layers with a bilinear pooling layer and a linear layer. Our MatConvNet-based~\cite{vedaldi15matconvnet} implementation runs between 30 to 100 frames per second on a NVIDIA Titan X GPU with cudnn-v5 depending on the model architecture. Even with faster object detection modules such as Faster R-CNNs~\cite{cortes2015nips} or Single-Shot Detector (SSD)~\cite{Liu2016eccv}, part-based models for fine-grained recognition are 2-10$\times$ slower. The main advantage of B-CNNs is that they require image labels only and can be easily applied to different fine-grained datasets.

\section{B-CNNs for Image Classification}\label{s:model}
In this section we introduce the B-CNN architecture for image classification and then show that various widely used texture representations can be written as B-CNNs.

\subsection{The B-CNN architecture}
A B-CNN for image classification consists of a quadruple ${\cal B} = (f_A, f_B, {\cal P}, {\cal C})$. Here $f_A$ and $f_B$ are \emph{feature functions} based on CNNs, ${\cal P}$ is a \emph{pooling function}, and ${\cal C}$ is a \emph{classification function}.  A feature function is a mapping $f:{\cal L}\times{\cal I} \rightarrow \real^{ K \times D}$, that takes an image $I \in {\cal I}$ and a location $l \in {\cal L}$ and outputs a feature of size $K \times D$. We refer to locations generally, which can include position and scale. The feature outputs are combined at each location using the matrix outer product, \ie, the bilinear combination of $f_A$ and $f_B$ at a location $l$ is given by 

\begin{equation} \label{eq:1}
\mathsf{bilinear}(l, I, f_A, f_B) = f_A(l, I)^Tf_B(l, I).
\end{equation}

Both $f_A$ and $f_B$ must have the same feature dimension $K$ to be compatible. The value of $K$ depends on the particular model. For example, $K=1$ for BoVW model and equals the number of clusters in a FV model (details in Section~\ref{s:related}). The pooling function ${\cal P}$ aggregates the bilinear combination of features across all locations in the image to obtain a global image representation $\Phi(I)$. We use sum pooling in all our experiments, \ie, 

\begin{equation} \label{eq:2}
\Phi(I) = \sum_{l \in {\cal L}} \mathsf{bilinear}(l,I, f_A, f_B) = \sum_{l \in {\cal L}} f_A(l, I)^Tf_B(l, I).
\end{equation}

\begin{figure}[t]
\begin{center}
\includegraphics[width=\linewidth]{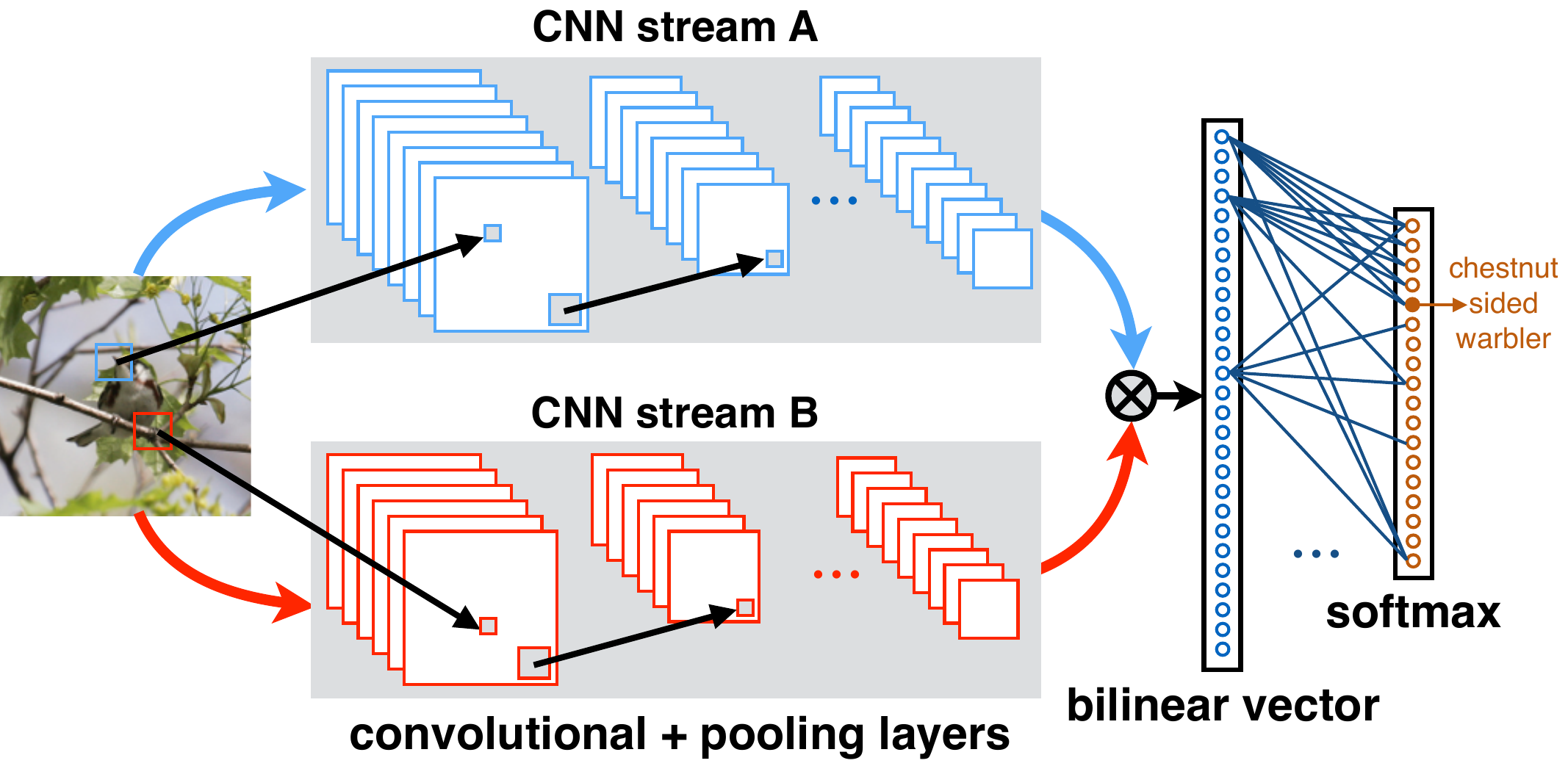}
\end{center}
\caption{\label{f:bcnn-architecture} \textbf{Image classification using a B-CNN.} An image is passed through CNNs A and B, and their outputs at each location are combined using the matrix outer product and average pooled to obtain the bilinear feature representation. This is passed through a linear and softmax layer to obtain class predictions.}
\end{figure}

Since the location of features is ignored during pooling, the bilinear feature $\Phi(I)$ is an \emph{orderless} representation. If $f_A$ and $f_B$ extract features of size $K\times M$ and $K \times N$ respectively, then $\Phi(I)$ is of size $M\times N$. The bilinear feature is a general-purpose image representation that can be used with a classifier ${\cal C}$ (Figure~\ref{f:bcnn-architecture}). Intuitively, the outer product conditions the outputs of features $f_A$ and $f_B$ on each other by considering their pairwise interactions, similar to the feature expansion in a quadratic kernel.

%Sec.~\ref{s:bcnn-training} describes a bilinear CNN model using deep networks as the feature functions and Sec.~\ref{s:training} shows how the bilinear form simplifies gradients, resulting in an end-to-end trainable network.
%Choosing different functions as $g(\mathbf{x})$ in this framework lets us represent various orderless texture descriptor models 
%as specific instances of the general bilinear model ${\cal B}$, (Sec.~\ref{s:related}). Further approximations to this representation gives us end-to-end trainable models as shown in Sec.~\ref{s:approximation}. 

\subsubsection{Feature functions}
\label{s:feature-functions} 
A natural candidate for the feature function $f$ is a CNN consisting of a hierarchy of convolutional and pooling layers. In our experiments we use CNNs pre-trained on the ImageNet dataset \emph{truncated} at an intermediate layer as feature functions. By pre-training we benefit when domain-specific data is limited. This has been shown to be effective for a number of tasks ranging from object detection, texture recognition, to fine-grained classification~\cite{donahue13decaf,girshick14rich,razavin14cnn-features,cimpoi14describing}. Another advantage of using CNNs is that the resulting network can process images of an arbitrary size and produce outputs indexed by image location and feature channel.

Our earlier work~\cite{lin2015bilinear} experimented with models where the feature functions $f_A$ and $f_B$ were either \emph{independent} or \emph{fully shared}. Here we also experiment with feature functions that share a part of the feed-forward computation as seen in Figure~\ref{fig:two_streams}. The feature functions used to approximate classical texture representations we present in Section~\ref{s:texture}, as well as the low-dimensional B-CNNs we present in Section~\ref{sec:dim_red}.

\subsubsection{Normalization and classification}
We perform additional normalization steps where the bilinear feature $\mathbf{x}=\Phi(I)$ is passed through a signed square-root ($\mathbf{y} \leftarrow \text{sign}(\mathbf{x}) \sqrt{|\mathbf{x}|}$), followed by $\ell_2$ normalization ($\mathbf{z} \leftarrow \mathbf{y} / ||\mathbf{y}||_2$) inspired by~\cite{perronnin10improving}. This improves performance in practice (see our earlier work~\cite{lin2015bilinear} for an evaluation of the effect of normalization). For classification we use logistic regression or linear SVM~\cite{scholkopf2001learning}. Although this can be replaced with an arbitrary multi-layer network, we found that linear models are effective on top of bilinear features.

\subsubsection{End-to-end training}\label{s:training}
Since the overall architecture is a directed acyclic graph, the parameters can be trained by back-propagating the gradients of the classification loss (\eg, cross-entropy). The \emph{bilinear form} simplifies the gradient computations. If the outputs of the two networks are matrices $A$  and  $B$ of size $L \times M$ and $L \times N$ respectively, then the bilinear feature is $\mathbf{x}=A^TB$ of size $M\times N$. Let ${d \ell}/{d\mathbf{x}}$ be the gradient of the loss function $\ell$ with respect to $\mathbf{x}$, then by chain rule of gradients we have:

\begin{equation}\label{eq:gradient}
	\frac{d \ell}{dA} = B \left( \frac{d \ell}{d\mathbf{x}} \right)^T , ~~~~\frac{d \ell}{dB} = A\left( \frac{d \ell}{d\mathbf{x}} \right).
\end{equation}

\begin{figure}[h]
\begin{center}
\includegraphics[width=\linewidth]{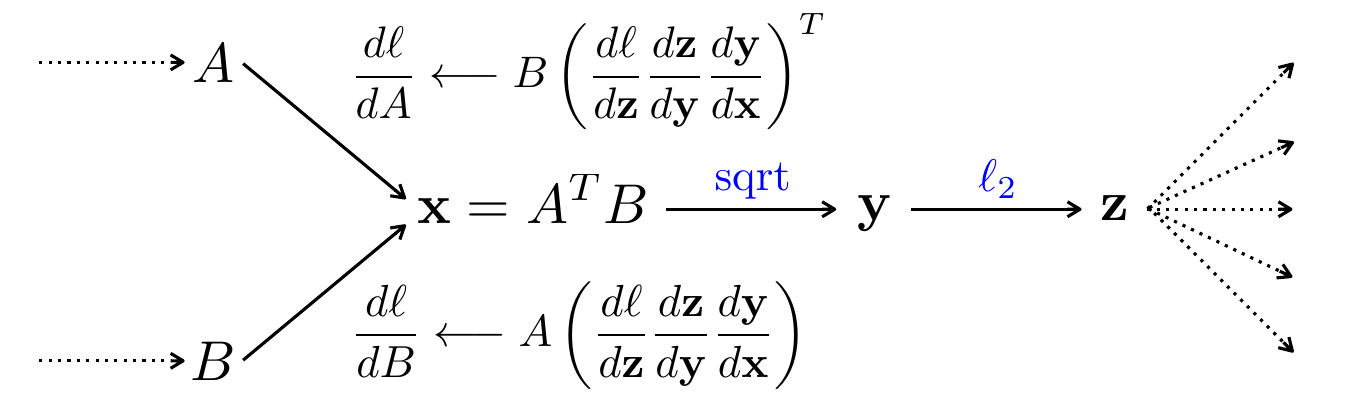}
\caption{\label{fig:gradients} Flow of gradients in a B-CNN.}
\end{center}
\end{figure}

As long as the gradients of the features $A$ and $B$ can be computed efficiently the entire model can be trained in an end-to-end manner. The scheme is illustrated in Figure~\ref{fig:gradients}.

\subsection{Relation to classical texture representations}\label{s:texture}
In this section we show that various \emph{orderless} texture descriptors can be written in the bilinear form and derive variants that are end-to-end trainable. Since the properties of texture are usually translationally invariant, most texture representations are based on orderless aggregation of local image features, \eg, sum or max operation.
A non-linear encoding is typically applied before aggregation of local features to improve their representation power. 
Additionally, a normalization of the aggregated feature (\eg, power and $\ell_2$) is done to increase invariance. 
Thus, texture representations can be defined by the choice of the \emph{local features}, the \emph{encoding function}, the \emph{pooling function}, and the \emph{normalization function}. 
To simplify further analysis, we will decompose the feature function $f$ as $f(l, \mathcal{I}) = g(h(l, \mathcal{I})) = g(\mathbf{x})$ to denote the explicit dependency on the image and the location of $h$ and additional non-linearities $g$.

One of the earliest representation used for texture is the Bag-of-Visual-Words (BoVW)~\cite{csurka04visual}. It was shown to be effective at several recognition tasks beyond texture. While variants differ on how the visual words are learned, a popular approach is to obtain a set of $k$ centers by clustering image features (\eg, using k-means). Each feature $\mathbf{x}$ is then assigned to the closest cluster center (also called ``hard assignment") and the image is represented as a histogram denoting frequencies of each visual word. If we denote $\eta(\mathbf{x})$ as the \emph{one-hot encoding} that is $1$ at the index of the closest center of $\mathbf{x}$ and zero elsewhere, then BoVW can be written as a bilinear model with $g_A(\mathbf{x}) = 1$ and $g_B(\mathbf{x}) = \eta(\mathbf{x})$. 

%. In this case the encoder assigns a feature $\mathbf{x}$ to the $k$ components based on its GMM posteriors. These encoded descriptors are averaged across the image to obtain the BoVW representation. Using our notation, if we set  $g_A(\mathbf{x})=1, \forall \mathbf{x}$, and $\eta(\mathbf{x})$ to be $g_B(\mathbf{x})$, the GMM posterior of the feature $\mathbf{x}$, the BoVW model can be written as a bilinear model $(1, \eta(\mathbf{x}), {\cal P}, {\cal C})$.

The VLAD representation\cite{jegou10aggregating} encodes a descriptor $\mathbf{x}$ as $(\mathbf{x} - \mu_k) \otimes \eta(\mathbf{x})$, where $\otimes$ is the \emph{kronecker product}, $\mu_k$ is the closest center to $\mathbf{x}$, and  $\eta(\mathbf{x})$ is the one-hot encoding of $\mathbf{x}$ as before.
%In the VLAD model, $\eta(\mathbf{x})$ is set to one for the closest center and zero elsewhere, also referred to as ``hard assignment." 
These encodings are aggregated across the image by sum pooling. Thus VLAD can be written as a bilinear model with $g_A(\mathbf{x}) = [\mathbf{x}-\mu_1; \mathbf{x}-\mu_2; \ldots; \mathbf{x}-\mu_k]$. Here, $g_A$ has $k$ rows each corresponding to a center. And $g_B(\mathbf{x}) = \text{diag}(\eta(\mathbf{x}))$, a matrix with $\eta(\mathbf{x})$ in the diagonal and $0$ elsewhere. Notice that the feature functions for VLAD output a matrix with $k > 1$ rows at each location. 

The FV representation~\cite{perronnin10improving} computes both the first order $\alpha_i = \Sigma_i^{-\frac{1}{2}}(\mathbf{x} - \mu_i)$ and second order $\beta_i=\Sigma_i^{-1}(\mathbf{x} - \mu_i)\odot (\mathbf{x} - \mu_i)-1$ statistics, which are aggregated and weighted by the Gaussian mixture model (GMM) posteriors $\theta(\mathbf{x})$. Here $\mu_i$ and $\Sigma_i$ are the mean and covariance of the $i^{th}$ GMM component respectively and $\odot$ denotes element-wise multiplication. Thus, FV can be written as a bilinear model with $g_A = [\alpha_1~\beta_1; \alpha_2 ~\beta_2;\ldots; \alpha_k~\beta_k]$ and $g_B = \text{diag}(\theta(\mathbf{x}))$.

The O2P representation~\cite{carreira2012semantic} computes the covariance statistics of SIFT features within a region, followed by log-Euclidean mapping and power normalization. Their approach was shown to be effective for semantic segmentation. O2P can can be written as a bilinear model with symmetric features, \ie, $f_A = f_B = \fsift$, followed by pooling and non-linearities.

The appearance-based cluster centers learned by the encoder, $\eta(\mathbf{x})$ or $\theta(\mathbf{x})$, in the BoVW, VLAD and FV representations can be thought of as part detectors. Indeed it has been observed that the cluster centers tend to localize facial landmarks when trained on faces~\cite{parkhi14a-compact}. Thus, by modeling the joint statistics of the encoder $\eta(\mathbf{x})$ or $\theta(\mathbf{x})$, and the appearance $\mathbf{x}$, the models can effectively describe appearance of parts regardless of where they appear in the image. This is particularly useful for fine-grained recognition where objects are not localized in the image.

%\begin{table}
%\begin{center}
%\begin{tabular}{|c|c|c|c|c}
%\hline 
%& \multicolumn{2}{|c|}{Exact} & \multicolumn{2}{|c|}{Approximation} \\
%\textbf{Model} & $g_A(\mathbf{x})$ & $g_B(\mathbf{x})$ & & \\ 
%\hline 
%O2P & $\mathbf{x}$ & $\mathbf{x}$ & & \\ 
%\hline 
%BOVW & $\text{diag}(\eta(\mathbf{x}))$ & $1$ & & \\ 
%\hline 
%VLAD & $\text{diag}(\eta(\mathbf{x}))$ & $ \mathbf{x} - \mathbf{C}$ & & \\ 
%\hline 
%FV & $\text{diag}(\eta(\mathbf{x}))$ & $[\alpha; \beta]$ & & \\  
%\hline 
%\end{tabular} 
%\end{center}
%\caption{\textbf{Orderless pooling descriptor models as bilinear models.} Various orderless pooling texture descriptor models can be written in the form of a bilinear model, $\cal B$, as described in Sec.~\ref{s:related}, by choosing appropriate functions $g_A$ and $g_B$ to act on the local descriptor $\mathbf{x}$.} \label{t:bilinear_approx}
%\end{table}

\begin{table*}
\caption{\label{t:bilinear_approx}
Texture encoders such as VLAD, FV, BoVW and O2P can be written as outer products of the form ${g_A}^T g_B$. On the right are their end-to-end trainable formulations that simplify gradient computations by replacing ``hard assignment" $\eta$ with ``soft assignment" $\bar{\eta}$, ignoring variance normalization for FV, \etc For the symmetric case (\ie, when $f_A = f_B$) bilinear pooling is identical to O2P. See Section~\ref{s:approximation} for details.}
\begin{center}
\renewcommand{\arraystretch}{1.5}
\begin{tabular}{c|c|c|c|c}
 & \multicolumn{2}{c}{\textbf{Exact formulation}} & \multicolumn{2}{|c}{\textbf{End-to-end trainable formulation}} \\
 \cline{2-5}
\textbf{Model} & $g_B(\mathbf{x})$ & $g_A(\mathbf{x})$ & $g_B(\mathbf{x})$ & $g_A(\mathbf{x})$ \\ 
\hline 
VLAD & $\text{diag}(\eta(\mathbf{x}))$ & $ \mathbf{x} - \boldsymbol{\mu}$ & $\text{diag}(\bar\eta(\mathbf{x}))$ & $ \mathbf{x} - \boldsymbol{\mu}$ \\ 
FV & $\text{diag}(\theta(\mathbf{x}))$ & $[\boldsymbol{\alpha}, \boldsymbol{\beta}]$ & $\text{diag}(\bar\eta(\mathbf{x}))$ & $[\mathbf{x}-\boldsymbol{\mu}, (\mathbf{x}-\boldsymbol{\mu}) \odot (\mathbf{x}-\boldsymbol{\mu})]$ \\ 
BoVW & $\eta(\mathbf{x})$ & $1$ & $\bar\eta(\mathbf{x})$ & $1$ \\ 
O2P &$\mathbf{x}$ & $\mathbf{x}$  & $\mathbf{x}$ & $\mathbf{x}$ \\
\end{tabular} 
\end{center}
\end{table*}

\begin{figure*}
\begin{center}
\begin{tabular}{@{}ccc@{}}
\includegraphics[height=0.125\linewidth]{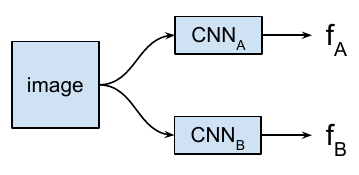} & 
\includegraphics[height=0.125\linewidth]{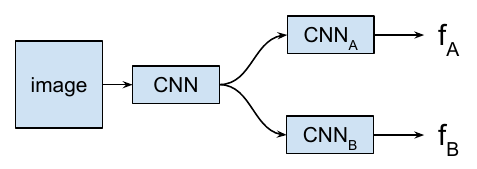} & 
\includegraphics[height=0.125\linewidth]{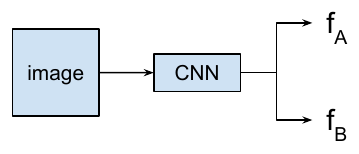} \\
(a) no sharing & (b) partially shared & (c) fully shared\\
\end{tabular}
\end{center}
\caption{\label{fig:two_streams} Feature functions in B-CNNs can (a) share no computations (\eg, B-CNN model based on VGG-M and VGG-D), (b) share computations partially (\eg, \netvlad, B-CNN PCA model described in Section~\ref{sec:dim_red}), and (c) share all computations (\eg, B-CNN model based on VGG-M).}
\end{figure*}

\subsubsection{End-to-end trainable formulations}\label{s:approximation}
Prior work on fine-grained recognition using texture encoders~\cite{cimpoi2016},~\cite{gosselin2014revisiting} did not learn the features in an end-to-end manner. Below we describe a recently proposed end-to-end trainable approximation of VLAD, and present similar formulations for all texture representations described in the earlier section. The ability to directly fine-tune these models leads to significant improvements in accuracy across a variety of fine-grained datasets.

%In this section we derive variants, and in some cases simplifying them, for which gradients have a simple form. These differentiable encoding layers can be plugged into existing deep architectures yielding a family of models that enjoy the benefits of features extracted from pre-trained models, the benefits offered by texture representations, as well as the ability to perform domain-specific fine-tuning. In our experiments we find that the ability to directly fine-tune these models leads to significant improvements in accuracy across a variety of fine-grained datasets.

Recently, an end-to-end trainable formulation of VLAD called NetVLAD was proposed by Arandjelovi\'c \etal~\cite{Arandjelovic16}. The first simplification was to replace the ``hard assignment" $\eta(\mathbf{x})$ in $g_B$ by a differentiable ``soft assignment" 
$\bar\eta(\mathbf{x})$. Given the $k$-th cluster center $\mu_k$, the $k$-th component of the soft assignment vector for an input $\mathbf{x}$ is given by,
\begin{equation}
\bar\eta_k(\mathbf{x}) = \frac{e^{- \gamma ||\mathbf{x} - \mu_k||^2}}{ \sum_{k'}e^{- \gamma ||\mathbf{x} - \mu_{k'}||^2} }
= \frac{e^{{\mathbf{w}_k}^T\mathbf{x} + b_k}}{ \sum_{k'}e^{{\mathbf{w}_{k'}}^T\mathbf{x} + b_{k'}} }
\end{equation}
where $\mathbf{w}_k = 2\gamma\mu_k$, $b_k = - \gamma ||\mu_k||^2$ and $\gamma$ is a parameter of the model. This is simply the \texttt{softmax} operation applied after a convolution layer 
with a bias term, and can be implemented using standard CNN building blocks.
The function $g_A$ remains unchanged
$[\mathbf{x}-\mu_1; \mathbf{x}-\mu_2; \ldots; \mathbf{x}-\mu_k]$. The second simplification is to decouple the dependence on $\mu$ of both the $g_A$ and $g_B$ during training which makes gradient computation easier. Thus in NetVLAD during training the weights $\mathbf{w}_k$, $b_k$ and $\mu_k$ are independent parameters.

We extend the NetVLAD to NetFV by appending the second order statistics to the feature $g_A $,
\ie,
$g_A = [\mathbf{x}-\mu_1, (\mathbf{x}-\mu_1)^2; \mathbf{x}-\mu_2, (\mathbf{x}-\mu_2)^2; \ldots; \mathbf{x}-\mu_k, (\mathbf{x}-\mu_k)^2 ]$.  Here, the squaring is done in an element-wise manner, \ie, 
$(\mathbf{x}-\mu_i)^2 = (\mathbf{x} - \mu_i)\odot (\mathbf{x} - \mu_i)$. The feature $g_B$ is kept identical to NetVLAD. 
This simplification discards the covariances and priors present in the 
true GMM posterior used in the FV model. 
Similarly, the \emph{NetBoVW} approximation to BoVW replaces the hard assignments by soft assignments $\bar\eta(\mathbf{x})$ computed in a manner similar to NetVLAD.

The O2P representation is identical to B-CNN when the feature functions $f_A$ and $f_B$ are identical. However, the O2P representation applies a log-Euclidean (matrix-logarithm) mapping to the pooled representation which is rather expensive to compute since it involves an Eigenvalue decomposition and currently does not have efficient implementation on GPUs. This significantly slows the forward and gradient computations of the entire network. Skipping this step allows us to efficiently fine-tune the model. We also note that concurrent to our publication~\cite{lin2015bilinear}, Ionescu \etal~\cite{ionescu2015matrix} proposed a DeepO2P approach and noted similar difficulties.

%In these approximations the centers $\boldsymbol{\mu}=[\mu_1, \mu_2, \ldots, \mu_k]$ can be initialized using GMMs or k-means. However, being trainable parameters these can be directly updated along with the rest of the network during training. Prior work has either held these fixed (\eg Fisher vector CNN~\cite{cimpoi14describing}), or resorted to indirect training by fine-tuning the standard CNN with fully-connected layers and then constructing Fisher vector representations on fine-tuned CNN features. Our experiments show a clear advantage of direct fine-tuning (Sec.~\ref{sec:exp_fg}).

\section{Image Classification Experiments}
\label{s:experiments}
We outline the models used in our experiments in Section~\ref{s:models}. We then provide a comparison of various B-CNNs to prior work on fine-grained recognition in Section~\ref{s:exp_fgvc}, and texture and scene recognition in Section~\ref{sec:exp_texture}.

\subsection{Models}\label{s:models}
Below we describe various models used in our experiments:

\textbf{FV with SIFT.} We implemented a FV based using dense SIFT features~\cite{perronnin10improving} extracted using VLFEAT~\cite{vedaldi2010vlfeat}. The image is first resized to 448$\times$448 and SIFT features with a \emph{bin size} of 8 pixels are computed densely across the image with a \emph{stride} of 4 pixels. The features are PCA projected to 80 dimensions before learning a GMM with 256 components.

\textbf{CNN with fully-connected (FC) layers.} This is a standard baseline where the features are extracted from the last FC layer, \ie, before the \texttt{softmax} layer of a CNN. The input image is resized to 224$\times$224 (the input size of the CNN) and mean-subtracted before propagating it though the CNN. We consider two different representations: 4096 dimensional \emph{relu7} layer outputs of both the VGG-M network~\cite{chatfield14return} and the 16-layer VGG-D network~\cite{simonyan14very}.

% For fine-tuning we replace the 1000-way classification layer trained on ImageNet dataset with a $k$-way softmax layer where $k$ is the number of classes in the fine-grained dataset. The parameters of the softmax layer are initialized randomly and we continue training the network on the dataset for several epochs at a smaller learning rate while monitoring the validation error. Once the networks are trained, the layer before the softmax layer is used to extract features.

\textbf{FV/NetFV with CNNs.} This denotes the method of~\cite{cimpoi2016} that builds a descriptor using FV pooling of CNN filter bank responses with 64 GMM components. One modification over~\cite{cimpoi2016} is that we first resize the image to 448$\times$448 pixels, \ie, twice the resolution the CNNs were trained on, and pool features from a \emph{single-scale}. This leads to a slight reduction in performance over the multi-scale approach. But we choose the single-scale setting because this keeps the feature extraction for all our methods identical making comparisons easier. We consider two representations based on the VGG-M and VGG-D networks. Unlike the earlier FC models, the features are extracted from the \emph{relu5} and \emph{relu5\_3} layers of the VGG-M and VGG-D networks respectively. 

\textbf{VLAD/NetVLAD with CNNs.} Similar to FV, this approach builds VLAD descriptors on CNN filter banks responses. We resize the image to 448$\times$448 pixels before passing it through the network and aggregate the features obtained from the CNN using VLAD/NetVLAD pooling with 64 cluster centers. Identical to the FV models, we consider VLAD models with VGG-M and VGG-D networks truncated at \emph{relu5} and \emph{relu5\_3} layers respectively.

\textbf{BoVW/NetBoVW with CNNs.} For BoVW we construct a vocabulary of 4096 words using k-means on top of the CNN features. For NetBoVW we use a 4096-way \texttt{softmax} layer as an approximation to the hard assignment. We use the same setting as FV and VLAD for feature extraction.

%\textbf{End-to-end texture models with CNN features} These models approximate classical texture representations as described in Sec.~\ref{s:approximation}. These encodings are denoted as \netbovw, \netvlad, and \netfv for approximations of BoVW, VLAD and Fisher vector respectively. Identical to the setting in \dcnn, the size of the input image is 448$\times$448 and an encoding layer with 64 components is used for \netvlad and \netfv representations, while a layer with 4096 components is used for the \netbovw representation (higher number of clusters resulted in a drop in accuracy, detailed in Sec.~\ref{sec:dim_red}). Some differences from the NetVLAD experiments in~\cite{Arandjelovic16} include: (1) we use the \emph{relu5}-layer features instead of \emph{conv5}-layer features and (2) we do not $\ell_2$-normalize the features. We found that normalization leads to a drop in performance in our experiments.

{\textbf{B-CNNs.} These are models presented in Section~\ref{s:model} where features from two CNNs are pooled using an outer product. When the two CNNs are identical the model is a extension of O2P using deep features without the log-Euclidean normalization. We consider several B-CNNs -- (i) one with two identical VGG-M networks truncated at the \emph{relu5} layer, (ii) one with a VGG-D and VGG-M network truncated at the \emph{relu5\_3} and \emph{relu5} layer respectively,  and (iii) initialized with two identical VGG-D networks truncated at the \emph{relu5\_3} layer. Identical to the setting in FV and VLAD, the input images are resized to 448$\times$448 and features are extracted using the two CNNs. The VGG-D network produces 28$\times$28 output compared to 27$\times$27 of the VGG-M network, so we downsample the VGG-D output by ignoring a row and column when combining it with the VGG-M output. The bilinear feature for all these models is of size 512$\times$512. Note that the symmetric models, \ie, option (i) and (iii), the two networks share all parameters (which is also the case when they are fine-tuned due to symmetry), so in practice these models have the same memory overhead and speed as a single network evaluation.

% which comparable to that of \dcnn (512$\times$128) and FV-SIFT (80 $\times$ 512). Although the representations for FV and VLAD have smaller dimensionality than the \bcnn feature, previous work~\cite{cimpoi2016} shows that these representations are effective in classification with 64 cluster centers. We empirically found that increasing the number of clusters did not show improvements in performance.
%For fine-tuning we add a $k$-way softmax layer. 
\vspace{-0.1in}
\subsubsection{Fine-tuning}
For fine-tuning we add $k$-way linear + softmax layer where $k$ is the number of classes in the fine-grained dataset. The parameters of the linear layer are initialized randomly. We adopt a two step training procedure of~\cite{branson14bird} where we first train the linear layer using logistic regression, a convex optimization problem, followed by fine-tuning the entire model using back-propagation for several epochs (about 45 -- 100 depending on the dataset and model) at a relatively small learning rate ($\eta=0.001$). Across the datasets we found the hyperparameters for fine-tuning were fairly consistent. Although, the exact VLAD, FV, BoVW models cannot be directly fine-tuned, we report results using \emph{indirect} fine-tuning where the networks are fine-tuned with FC layers.  We found this improves accuracy. For example, the indirectly fine-tuned FV models with CNN features outperforms the multi-scale but not fine-tuned results reported in~\cite{cimpoi2016}. However, direct fine-tuning using NetFV is significantly better.

\subsubsection{SVM training and evaluation}
In all our experiments before and after fine-tuning, training and validation sets are combined and one-vs-all linear SVMs on the extracted features are trained by setting the learning hyperparameter $C_{\text{svm}}=1$. Since our features are $\ell_2$ normalized, the optimal of $C_{\text{svm}}$ is likely to be independent of the dataset. The trained classifiers are calibrated by scaling the weight vector such that the median scores of positive and negative training examples are at $+1$ and $-1$ respectively. For each dataset we double the training data by flipping images and at test time average the predictions of the image and its flipped copy. SVM training provides 1-3\% improvement over logistic regression with the VGG-M networks, but provides \emph{negligible} improvement with the VGG-D networks. 
Test time flipping improves performance by 0.5\% on average for the VGG-M networks, while has \emph{negligible} impact on the accuracy for the VGG-D networks. 
Performance is measured as the percentage of correctly classified images for all datasets.

%\subsubsection{Main conclusions} To stress the points we learn from the experiments, we summarize the experiments section and draw the main conclusion as follow:
%\begin{enumerate}
%\item Across all fine-grained datasets we use, orderless pooling models like VLAD, FV, and O2P work better than fully connected pooling.
%\item The approximation of VLAD and FV using bilinear models achieve comparable results to the standard models and allow end-to-end fine-tuning.
%\item The direct fine-tuning of bilinear approximation of VLAD and FV significantly improves the accuracy than indirect fine-tuning of original models via fine-tuning corresponding \rcnn.
%\item Bilinear O2P models outperform other pooling schemes on most cases.
%\end{enumerate}

\subsection{Fine-grained recognition}\label{s:exp_fgvc}

We evaluate methods on following fine-grained datasets and report the per-image accuracy in Table~\ref{tab:results}.

\textbf{CUB-200-2011}~\cite{WahCUB_200_2011} dataset contains 11,788 images of 200 bird species which are split into roughly equal train and test sets with detail annotation of parts and bounding boxes. As birds appear in different poses and viewpoints and occupy small portion of image in cluttered background, classifying bird species is challenging. Notice that in all our experiments, we only use image labels during training without any part or bounding box annotation. In the following sections, "birds" refers to the results on this dataset.

\textbf{FGVC-aircraft} dataset~\cite{maji2013fine} consists of 10,000 images of 100 aircraft variants, and was introduced as a part of the FGComp 2013 challenge.  The task involves discriminating variants such as the Boeing 737-300 from Boeing 737-400. The differences are subtle, \eg, one may be able to distinguish them by counting the number of windows in the model. Unlike birds, airplanes tend to occupy a significantly larger portion of the image and appear in relatively clear background. Airplanes also have a smaller representation in the ImageNet dataset compared to birds.

\textbf{Stanford cars} dataset~\cite{krause20133d} contains 16,185 images of 196 classes. Categories are typically at the level of Make, Model, Year, \eg, ``2012 Tesla Model S" or `2012 BMW M3 coupe." Compared to aircrafts, cars are smaller and appear in a more cluttered background. Thus object and part localization may play a more significant role here. This dataset was also part of the FGComp 2013 challenge.

\textbf{NABirds}~\cite{van2015building} is larger than the CUB dataset consisting of 48,562 images of 555 spices of birds that include most that are found in North America. The work engaged citizen scientists to produce high-quality annotations in a cost-effective manner. This dataset also provides parts and bounding-box annotations, but we only use category labels for training our models.  

\subsubsection{Bird species classification}\label{s:bird-results}
%Several methods report results requiring varying degrees of supervision such as part annotation or bounding-boxes at training and test time. We refer readers to~\cite{branson14bird} that has a comprehensive discussion of results on this dataset. A more up-to-date set of results can be found in~\cite{krause2015fine} who recently reported excellent performance on this dataset leveraging more accurate CNN models with a method to train part detectors in a weakly supervised manner.

\textbf{Comparison to baselines.} Table~\ref{tab:results} ``bird" column shows results on the CUB-200-2011 dataset. 
The end-to-end approximations of texture representations (NetBoWV, NetVLAD, NetFV) improve significantly after fine-tuning. Exact models with indirect fine-tuning (\ie, fine-tuned with FC layers) also improve, but the improvement is smaller (shown in gray italics in Table~\ref{tab:results}).
With fine-tuning the single-scale FV models outperforms the multi-scale results reported in~\cite{cimpoi2016} -- 49.9\% using VGG-M and 66.7\% using VGG-D network.
B-CNNs offer the best accuracy across all models with the best performing model obtaining \textbf{84.1\%} accuracy (VGG-M + VGG-D). The next best approach is the NetVLAD with 81.9\% accuracy. We found that increasing the cluster centers does not improve performance of NetVLAD (see Section~\ref{sec:dim_red}).

We also trained the B-CNN (VGG-M + VGG-D) model on the much larger \emph{NABirds dataset}. For this model we skipped the SVM training step and report the accuracy using \texttt{softmax} layer predictions. This model achieves \textbf{79.4\%} accuracy outperforming a fine-tuned VGG-D network that obtains 63.7\% accuracy. Van Horn~\etal~\cite{van2015building} obtains \textbf{75\%} accuracy using AlexNet and part annotations at test time, while the ``neural activation constellations" approach \cite{simon2015neural} obtains \textbf{76.3\%} accuracy using a GoogLeNet architecture~\cite{Szegedy_2015_CVPR}.

\begin{table*}
\caption{Per-image accuracy on the birds~\cite{WahCUB_200_2011}, aircrafts~\cite{maji2013fine}, and cars~\cite{krause20133d} datasets for various methods described in Section~\ref{s:models}. We compare various texture representations and prior work (separated by a double line). The texture representations the first column lists the features used in the encoding followed by the pooling strategy. Features are extracted from the \emph{relu5} and \emph{relu5\_3} layer of VGG-M and VGG-D networks respectively. FC pooling corresponds to fully-connected layers on top these intermediate layer features such that it corresponds to the penultimate layer of the original network. The second and third columns show additional annotations used during training and testing. Results are shown without and with domain-specific fine-tuning. Directly fine-tuning the approximate models leads to  better performance than indirectly fine-tuning (shown in gray italics). Features are constructed from the corresponding layers of the fine-tuned FC-CNN models. B-CNN models achieve the best accuracy across texture representations. The first, second, and third best texture models are marked with red, blue and yellow colors respectively.}
\setlength{\tabcolsep}{8pt}
\renewcommand{\arraystretch}{1.5}
\begin{center}
\begin{tabular}{llll|cc|cc|cc}
 & & & & \multicolumn{2}{c|}{\bf{birds}}  &  \multicolumn{2}{c|}{\bf{aircrafts}} & \multicolumn{2}{c}{\bf{cars}} \\ 
\cline{5-10} 
\textbf{features} & \textbf{train} & \textbf{test} & \textbf{encoding} & w/o ft & w/ ft &w/o ft & w/ ft & w/o ft & w/ ft \\ 
\hline 
SIFT & & & FV & 18.8   & - & 61.0  & - & 59.2& - \\ 
\hline
& & & FC & 52.7  & 58.8 & 44.4  & 63.4 & 37.3& 58.6 \\ 
& & & BoVW & 41.9 & \textcolor{Gray}{\textit{43.8}} & 56.2 & \textcolor{Gray}{\textit{60.1}} & 54.2 & \textcolor{Gray}{\textit{58.4}} \\ 
& & & \netbovw & 47.9 & 48.6 & 58.8 & 65.9 & 60.3 & 66.1 \\ 
& & & VLAD & 66.5 & \textcolor{Gray}{\textit{70.5}} & 70.5 & \textcolor{Gray}{\textit{74.8}} & 75.3 & \textcolor{Gray}{\textit{78.9}} \\ 
VGG-M (\emph{relu5}) & \na & \na & \netvlad & 66.8 & 72.1 & 70.7 & 76.7 & 76.0 & 83.7 \\ 
& & & FV & 61.1  & \textcolor{Gray}{\textit{64.1}} & 64.3  & \textcolor{Gray}{\textit{71.2}} & 70.8  & \textcolor{Gray}{\textit{77.2}} \\
& & & \netfv & 64.5 & 71.7 & 68.6 & 75.5 & 72.3 & 81.8 \\
& & & \bcnn & 72.0 & 78.1 & 72.7  & 79.5  & 77.8 & 86.5 \\
\hline 
& & & FC & 61.0 & 70.4 & 45.0  & 76.6  & 36.5  & 79.8 \\
& & & BoVW & 56.6 & \textcolor{Gray}{\textit{58.8}} & 61.9  & \textcolor{Gray}{\textit{71.3}} & 62.7 & \textcolor{Gray}{\textit{73.9}} \\ 
& & & \netbovw & 65.9 & 69.7 & 65.1 & 74.0 & 71.0 & 76.7 \\ 
& & & VLAD & 78.0 & \textcolor{Gray}{\textit{79.0}} & 75.2 & \textcolor{Gray}{\textit{80.6}} & 81.9 & \textcolor{Gray}{\textit{85.6}} \\ 
VGG-D (\emph{relu5\_3}) & \na & \na & \netvlad & 77.9 & \cellcolor{lightyellow}81.9 & 75.3 & \cellcolor{lightyellow}81.8 & 82.1 & \cellcolor{lightyellow}88.6 \\ 
& & & FV & 71.3  & \textcolor{Gray}{\textit{74.7}} & 70.4 & \textcolor{Gray}{\textit{78.7}}  & 75.2 & \textcolor{Gray}{\textit{85.7}} \\  
& & & \netfv & 73.9 & 79.9 & 71.5 & 79.0 & 77.9 & 86.2 \\
& & & \bcnn & 80.1 & \cellcolor{lightblue}84.0 & 77.7 &  \cellcolor{lightred}86.9 & 82.9 & \cellcolor{lightblue}90.6 \\ 
\hline

VGG-M + VGG-D & & &\bcnn & 80.1 & \cellcolor{lightred}84.1 & 78.0 & \cellcolor{lightblue}86.6 & 83.9 & \cellcolor{lightred}91.3 \\ 
\hline
\hline 
VGG-D & \na & \na & PD+SWFV-CNN~\cite{Zhang2016CVPRpdfs} & \multicolumn{2}{c|}{83.6} & \multicolumn{2}{c|}{-} & \multicolumn{2}{c}{-} \\
VGG-D & \na & \na & PD+FC+SWFV-CNN~\cite{Zhang2016CVPRpdfs} & \multicolumn{2}{c|}{84.5} & \multicolumn{2}{c|}{-} & \multicolumn{2}{c}{-} \\
Inception-BN & \na & \na & STNs~\cite{jaderberg15spatial} & \multicolumn{2}{c|}{84.1} & \multicolumn{2}{c|}{-} & \multicolumn{2}{c}{-} \\
VGG-D & Box & \na & Krause \etal~\cite{krause2015fine} & \multicolumn{2}{c|}{82.0} & \multicolumn{2}{c|}{-} & \multicolumn{2}{c}{\textbf{92.6}} \\
VGG-D & Box+Part & Box & SPDA-CNN~\cite{Zhang2016CVPRspda} & \multicolumn{2}{c|}{84.6} & \multicolumn{2}{c|}{-} & \multicolumn{2}{c}{-} \\
VGG-D & Box & Box & CrossLayerPooling~\cite{liu2016cross} & \multicolumn{2}{c|}{77.0} & \multicolumn{2}{c|}{-} & \multicolumn{2}{c}{-} \\
VGG-D & Part & \na & Zhang \etal~\cite{zhangiclr2016} & \multicolumn{2}{c|}{85.9} & \multicolumn{2}{c|}{-} & \multicolumn{2}{c}{-} \\
AlexNet & Box+Part & \na & Part-based RCNN~\cite{zhang14part-based} & \multicolumn{2}{c|}{73.9} & \multicolumn{2}{c|}{-} & \multicolumn{2}{c}{-} \\
AlexNet & Box+Part & \na & Branson \etal~\cite{branson14bird} & \multicolumn{2}{c|}{75.7} & \multicolumn{2}{c|}{-} & \multicolumn{2}{c}{-} \\
B-CNN (VGG-D) & \na & \na & BoostCNN~\cite{MoghimiBMVC2016} & \multicolumn{2}{c|}{\textbf{86.2}} & \multicolumn{2}{c|}{\textbf{88.5}} & \multicolumn{2}{c}{92.1} \\
VGG-D & Box & Box & BoT~\cite{Wang2016CVPRbot} & \multicolumn{2}{c|}{-} & \multicolumn{2}{c|}{88.4} & \multicolumn{2}{c}{92.5} \\
FV+SIFT & \na & \na & Gosselin \etal~\cite{gosselin2014revisiting} & \multicolumn{2}{c|}{-} & \multicolumn{2}{c|}{80.7} & \multicolumn{2}{c}{82.7} \\
%Color+SIFT & Box & \na & Symbiotic~\cite{chai2013symbiotic} & \multicolumn{2}{c|}{59.4 (per-class)} & \multicolumn{2}{c|}{72.5} & \multicolumn{2}{c}{78.0} \\
\hline
\end{tabular}
\end{center}
\label{tab:results}
\vspace{-0.1in}
\end{table*}

\textbf{Comparison to other techniques.} Table~\ref{tab:results} shows other top-performing methods on this dataset. The dataset also provides bounding-box and part annotations and techniques differ based on what annotations are used at training and test time (also shown in the Table). Two early methods that performed well when bounding-boxes are not available at test time are 73.9\% of the ``part-based R-CNN"~\cite{zhang14part-based} and 75.7\% of the ``pose-normalized CNN"~\cite{branson14bird}. These methods are based on AlexNet~\cite{krizhevsky12imagenet} and can be improved with deeper and more accurate networks such as the VGG-D. For example, the SPDA-CNN~\cite{Zhang2016CVPRspda} trains better part detectors and feature representations using the VGG-D network and report 84.6\% accuracy. Krause \etal~\cite{krause2015fine} report 82.0\% accuracy using a weakly-supervised method to learn part detectors, followed by the part-based analysis of~\cite{zhang14part-based} using the VGG-D network. However, our approach is simpler and faster since it does not rely on training and evaluating part detectors. The ``cross-layer pooling" technique~\cite{liu2016cross} that considers pairwise features extracted from two different layers of a CNN reports an accuracy of 73.5\% using AlexNet and 77.0\% accuracy with VGG-D. The approach uses bounding-boxes during training and testing.

One of the top-performing approaches that does not rely on additional annotations is the Spatial Transformer Networks~\cite{jaderberg15spatial}. It obtains \textbf{84.1\%} accuracy using the batch-normalized Inception network~\cite{ioffe2015batch}. The PD+SWFV-CNN approach combines unsupervised part detection with FV pooling of CNN features to obtain 83.6\% accuracy, and \textbf{84.5\%} accuracy with FC and FV pooling.

Since its publication, the B-CNN model has been improved by others in several ways. First, the \emph{compact bilinear pooling} approach~\cite{Gao2016CVPR} was proposed to reduce the size of the bilinear features (we evaluate this in Section~\ref{sec:dim_red}). Zhang \etal~\cite{zhangiclr2016} combine B-CNNs with part annotations and improve results to \textbf{85.9\%}. Moghimi \etal~\cite{MoghimiBMVC2016} boost B-CNNs trained on images of varying resolutions to obtain \textbf{86.2\%} accuracy. Although not directly comparable, Krause \etal ~\cite{krause2016unreasonable} show that by training on two orders of magnitude more labeled data obtained by querying category labels on search engines, the performance of deep architectures can be improved to 92.1\%.

\textbf{Common mistakes.} Figure~\ref{fig:cub-confusion} shows the top six pairs of classes that are confused by our fine-tuned B-CNN (VGG-M + VGG-D) model. The most confused pair of classes is ``American crow"  and ``Common raven". The differences lie in the wing-spans, habitat, and voice, none of which are easy to measure from the image. Other confused classes are also similar -- various Shrikes, Terns, Flycatchers, Cormorants and Gulls. We note that the dataset has an estimated 4.4\% label noise hence some of these errors may come from incorrect labeling~\cite{van2015building}.

\begin{figure}
\begin{center}
\begin{tabular}{@{}c|c@{}}
\includegraphics[width=0.45\linewidth]{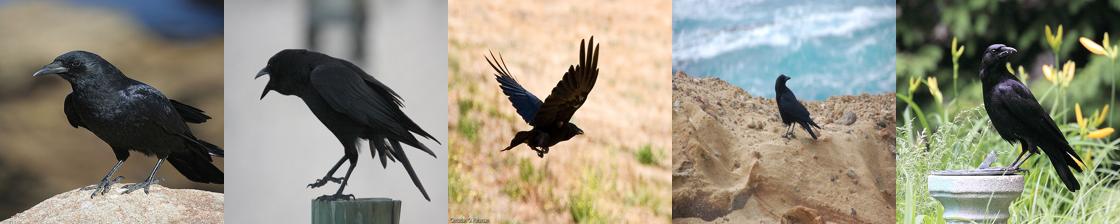} &
\includegraphics[width=0.45\linewidth]{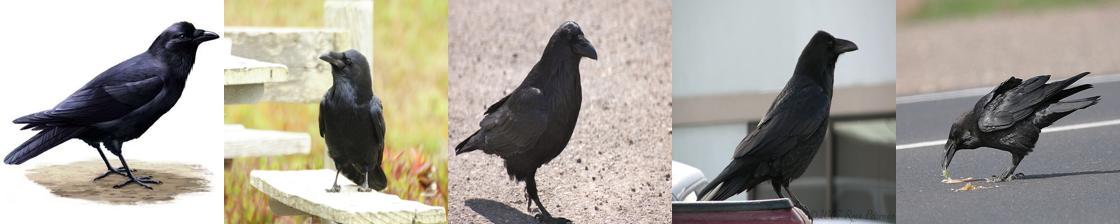} \\
American\_Crow &
Common\_Raven \\
\includegraphics[width=0.45\linewidth]{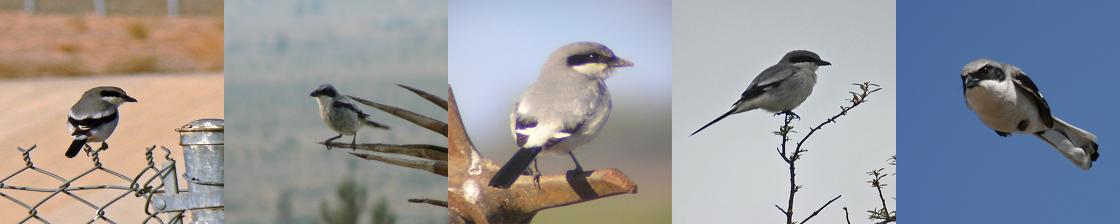} &
\includegraphics[width=0.45\linewidth]{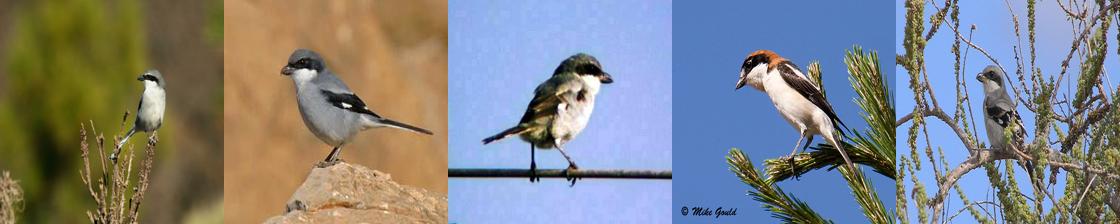} \\
Loggerhead\_Shrike &
Great\_Grey\_Shrike \\
\includegraphics[width=0.45\linewidth]{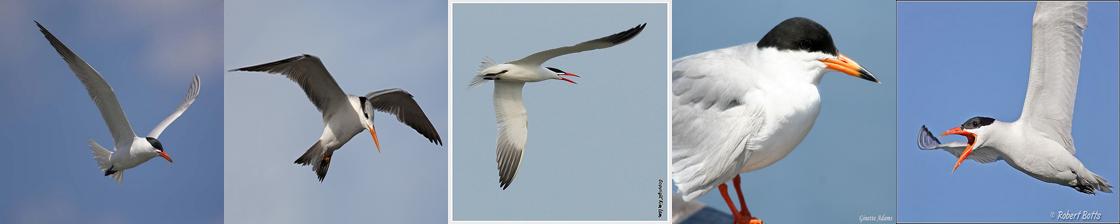} &
\includegraphics[width=0.45\linewidth]{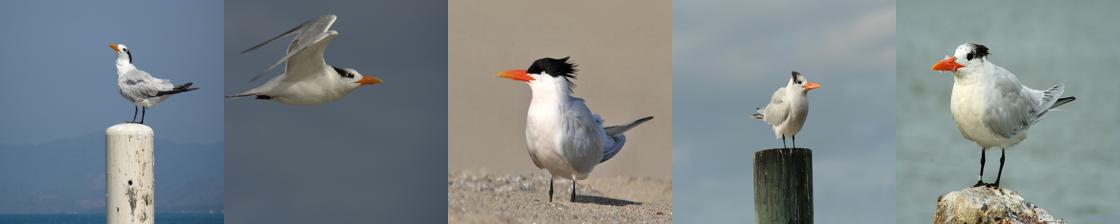} \\
Caspian\_Tern &
Elegant\_Tern \\
\includegraphics[width=0.45\linewidth]{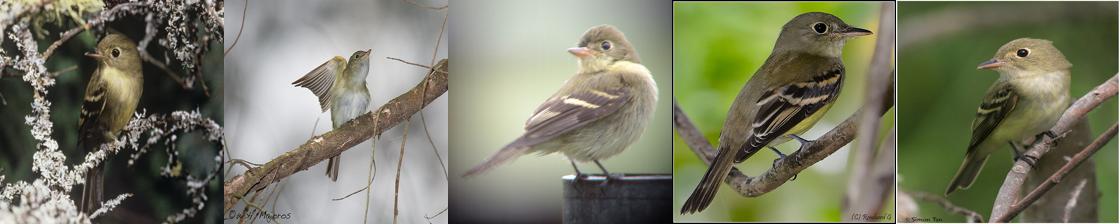} &
\includegraphics[width=0.45\linewidth]{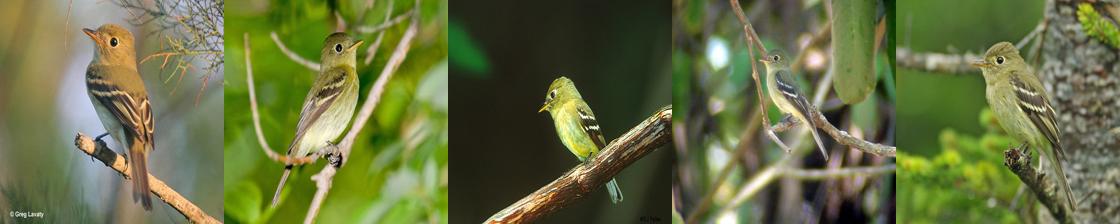} \\
Acadian\_Flycatcher &
Yellow\_bellied\_Flycatcher \\
\includegraphics[width=0.45\linewidth]{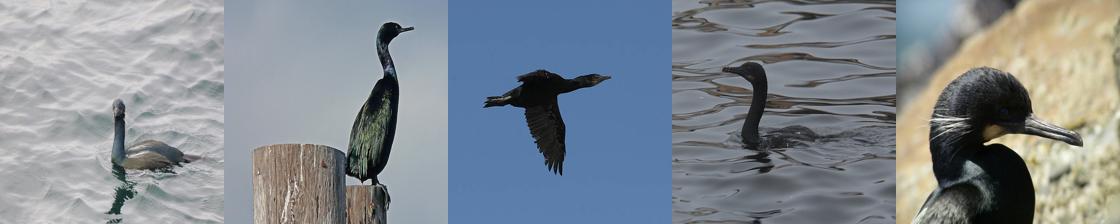} &
\includegraphics[width=0.45\linewidth]{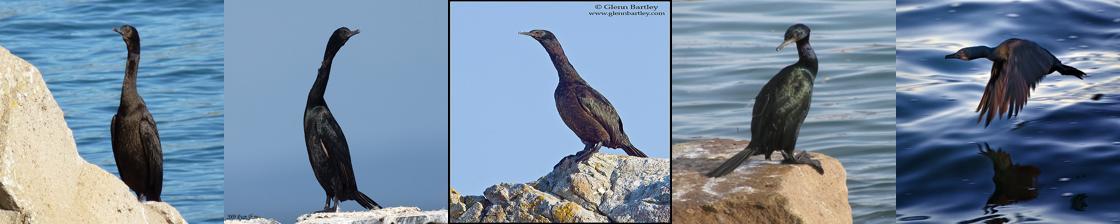} \\
Brandt\_Cormorant &
Pelagic\_Cormorant \\
\includegraphics[width=0.45\linewidth]{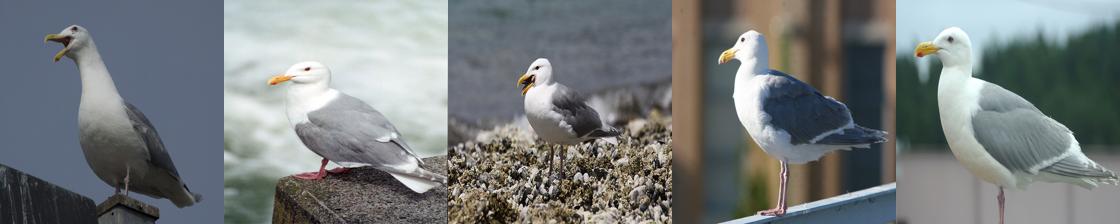} &
\includegraphics[width=0.45\linewidth]{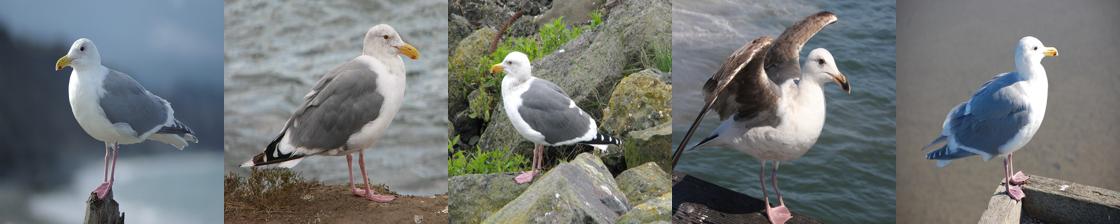} \\
Glaucous\_winged\_Gull &
Western\_Gull \\
\end{tabular}
\end{center}
\caption{\label{fig:cub-confusion} Top six pairs of classes that are most confused with each other on the CUB dataset. In each row we show the images in the test set that were most confidently classified as the class in the other column.}
\end{figure}  

\subsubsection{Aircraft variant classification}
\textbf{Comparison to baselines.} The trends among the baselines are similar to those in birds with a few exceptions. The FV with SIFT is remarkably good (61.0\%) and comparable to some of the CNN baselines. Compared to the birds, the effect of fine-tuning is significantly larger for models based on the VGG-D network suggesting a larger domain shift from the ImageNet dataset. As aircrafts appear mostly on the image center, cropping the central image improves the accuracy over our earlier work~\cite{lin2015bilinear}. We resize the images into $512 \times 512$ and then crop the central $448 \times 448$ as input. This achieves the best performance of \textbf{86.9\%} by the \bcnn (VDD-D) model. NetVLAD obtains 81.4\% accuracy.

\textbf{Comparison to other techniques.} This dataset does not come with part annotations hence several top performing methods for the birds dataset are not applicable here. We also compare against the results for ``track 2", \ie, w/o bounding-boxes, at the FGComp 2013 challenge~\cite{FGComp13}. The best performing method~\cite{gosselin2014revisiting} is a heavily engineered FV-SIFT which achieves 80.7\% accuracy. Notable differences between our baseline FV-SIFT and theirs are (i) larger dictionary (256 $\rightarrow$ 1024), (ii) Spatial pyramid pooling (1$\times$1 $\rightarrow$ 1$\times$1 + 3$\times$1), (iii) multiple SIFT variants, and (iv) multi-scale SIFT. Wang \etal~\cite{Wang2016CVPRbot} report \textbf{88.4\%} accuracy by mining discriminative patch triplets, but require bounding boxes during training and testing. Boosting B-CNNs~\cite{MoghimiBMVC2016} obtains the current state of the art with \textbf{88.5\%} accuracy.

%The next best method is the ``symbiotic segmentation" approach of~\cite{chai2013symbiotic} that achieves 72.5\% accuracy. However, this method requires bounding-box annotations at training time to learn a detector which is refined to a foreground mask. The \bcnn models outperform these methods by a significant margin. The results on this dataset show that orderless pooling methods are still of considerable importance -- they can be easily applied to new datasets as they only need image labels for training.

\subsubsection{Car model classification} 

\textbf{Comparison to baselines.} FV with SIFT does well on this dataset achieving 59.2\% accuracy. The effect of fine-tuning on cars is larger in comparison to birds and airplanes. Once again the B-CNNs outperform all the other baselines with the \bcnn (VGG-D + VGG-M) model achieving \textbf{91.3\%} accuracy. NetVLAD obtains \textbf{88.6\%} accuracy.

\textbf{Comparison to other techniques.} The best accuracy on this dataset is by Krause \etal~\cite{krause2015fine} which obtains  
92.6\% accuracy. This is closely matched by \textbf{92.5\%} accuracy of the discriminative patch triplets~\cite{Wang2016CVPRbot}, and \textbf{92.1\%} of Boosted B-CNNs~\cite{MoghimiBMVC2016}. Unlike the other approaches, Boosted B-CNNs do not rely on bounding-boxes at training time.

%The FV with SIFT ensemble~\cite{gosselin2014revisiting} obtains remarkable 82.7\% accuracy. The symbiotic segmentation achieved 78.0\% accuracy. The fine-tuned \bcnn[D,M] obtains 91.3\% significantly outperforming the SIFT ensemble, and nearly matching~\cite{krause2015fine} which requires bounding-boxes during training. The results when bounding-boxes are available at test time can be seen in ``track 1" of the FGComp 2013 challenge and are also summarized in~\cite{gosselin2014revisiting}. The SIFT ensemble improves significantly with the addition of bounding-boxes (82.7\% $\rightarrow$ 87.9\%) in the cars dataset compared to aircraft dataset where it improves marginally (80.7\% $\rightarrow$ 81.5\%). This shows that localization in the cars dataset is more important than in aircrafts. Our bilinear models have a clear advantage over FV models in this setting since it can learn to ignore the background clutter.

\subsection{Texture and scene recognition}
\label{sec:exp_texture}

We experiment on three texture datasets -- the \emph{Describable Texture Dataset}~(DTD)~\cite{cimpoi14describing}, \textit{Flickr Material Dataset} (FMD)~\cite{sharan09material}, and \emph{KTH-TISP2-b} (KTH-T2b)~\cite{caputo05class}. DTD consists of 5640 images labeled with 47 describable texture attributes.  FMD consists of 10 material categories, each of which contains 100 images. Unlike DTD and FMD where images are collected from the Internet, KTH-T2b contains 4752 images of 11 materials captured under controlled scale, pose, and illumination. The KTH-T2b dataset splits the images into four samples for each category. We follow the standard protocol by training on one sample and test on the remaining three. On DTD and FMD, we randomly divide the dataset into 10 splits and report the mean accuracy across splits. Besides these, we also evaluate our models on \textit{MIT indoor scene} dataset~\cite{quattoni09recognizing}. Indoor scenes are weakly structured and orderless texture representations have been shown to be effective here. The dataset consists of 67 indoor categories and a defined training and test split.

We compare \bcnn to the prior state-of-the-art approach of FV pooling of CNN features~\cite{cimpoi2016} using the VGG-D network. These results are without fine-tuning. On the MIT indoor dataset fine-tuning B-CNNs leads to a small improvement $72.8\% \rightarrow 73.8\%$ using \emph{relu5\_3} at $s=1$, while on the other datasets the improvements were negligible, likely due to the relatively small size of these datasets. Table~\ref{tab:texture_scale} shows the results obtained by features from a single scale and features from multiple scales $2^s, s\in$ \{1.5:-0.5:-3\} relative to the 224$\times$224 image using B-CNN and FV representations. We discard scales for which the image is smaller than the size of the receptive fields of the filters, or larger than $1024^2$ pixels for efficiency. Across all scales of the input image the performance of the two approaches are identical. Multiple scales consistently lead to an improvement in accuracy. The multi-scale FV results reported here are comparable ($\pm1\%$) to the results reported in Cimpoi \etal~\cite{cimpoi2016} for all datasets except KTH-T2b ($-4\%$). These differences are due to the choice of the CNN (they use the \emph{conv5\_4} layer of the 19-layer VGG network) and the range of scales. These results show that the B-CNNs are comparable to the FV pooling for texture recognition. One drawback is that the FV features with 64 GMM components has smaller in size (64$\times$2$\times$512) than the bilinear features (512$\times$512). However, bilinear features are highly redundant and their dimensionality can be reduced by an order of magnitude without loss in performance (see Section~\ref{sec:dim_red}).

\begin{table}
\caption{\label{tab:texture_scale} Mean per-class accuracy on DTD, FMD, KTH-T2b and MIT indoor datasets using FV and B-CNN representations constructed on top of \emph{relu5\_3} layer outputs of the 16-layer VGG-D network~\cite{simonyan14very}. Results for input images at different scales $s=1$, $s=2$ and $ms$ correspond to a size of 224$\times$224, 448$\times$448 and multiple sizes respectively.}
\vspace{-0.15in}
\small
\renewcommand{\arraystretch}{1.3}
\setlength{\tabcolsep}{4pt}
\begin{center}
\begin{tabular}{l|ccc|ccc}
 & \multicolumn{3}{c|}{\bf{FV}} & \multicolumn{3}{c}{\bf{\bcnn}}  \\ 
\cline{2-7} 
\textbf{dataset} & $s=1$ & $s=2$ & $ms$ & $s=1$ & $s=2$ & $ms$ \\ 
\hline 
DTD & $67.8$  & $70.6$ & $73.6$  & $69.6$ & $71.5$ & $72.9$\\ 
        & $^{\pm0.9}$   & $^{\pm0.9}$ & $^{\pm1.0}$  & $^{\pm0.7}$ & $^{\pm0.8}$ & $^{\pm0.8}$\\ 
FMD & $75.1$  & $79.0$  & $80.8$ & $77.8$ & $80.7$  & $81.6$ \\ 
 & $^{\pm2.3}$  & $^{\pm1.4}$  & $^{\pm1.7}$ & $^{\pm1.9}$ & $^{\pm1.5}$  & $^{\pm1.7}$ \\ 
KTH-T2b & $74.8$  & $75.9$ & $77.9$ & $75.1$ & $76.4$ & $77.9$ \\ 
 & $^{\pm2.6}$  & $^{\pm2.4}$ & $^{\pm2.0}$ & $^{\pm2.8}$ & $^{\pm3.5}$ & $^{\pm3.1}$ \\ 
MIT indoor& $70.1$ & $78.2$  & $78.5$  & $72.8$ & $77.6$  & $79.0$ \\ 
\end{tabular}
\end{center}
\vspace{-0.2in}
\end{table}

\section{Analysis of bilinear CNNs}\label{sec:analysis}

\subsection{Dimensionality reduction}
\label{sec:dim_red}
The outer product of CNN features generates very high dimensional image descriptors, \eg, 262K for the \bcnn models in Table~\ref{tab:results}. 
Our earlier work~\cite{Lin2016CVPR} showed that the features are highly redundant and their dimensionality can be reduced by an order of magnitude without loss in classification performance. Prior work~\cite{jegou10aggregating} has also shown that in the context of SIFT-based FV and VLAD, highly compact representations can be obtained.

\begin{figure*}[!htbp]
\begin{center}
\begin{tabular}{@{}ccc@{}}
\includegraphics[width=0.32\linewidth]{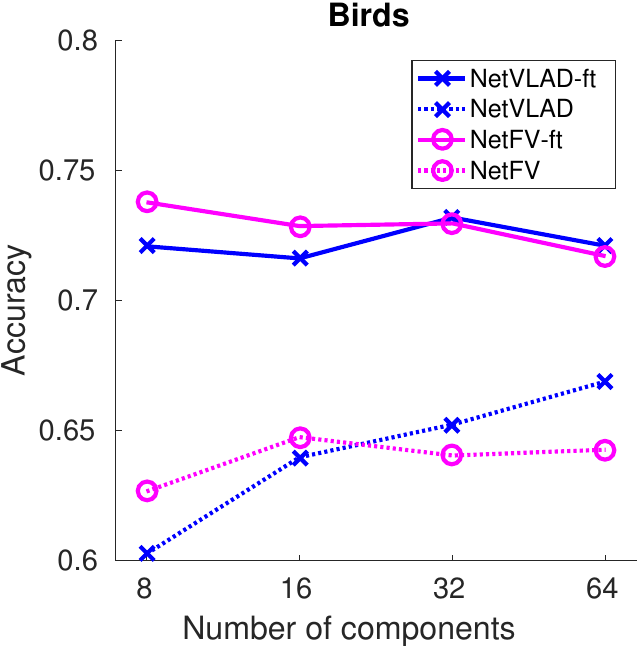} &
\includegraphics[width=0.32\linewidth]{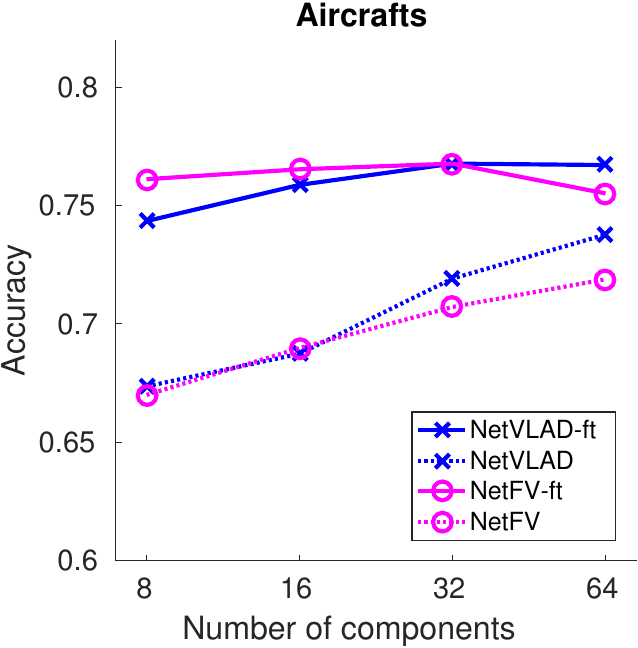} &
\includegraphics[width=0.32\linewidth]{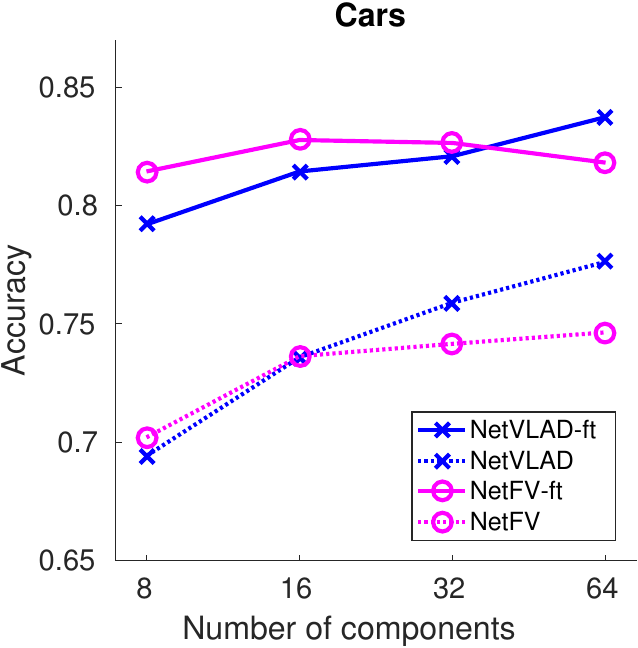} \\
\end{tabular}
\end{center}
\caption{\label{fig:gmm_components} Performance of \netvlad and \netfv models encoding VGG-M \textit{relu5} features with different number of cluster centers on fine-grained datasets before (dashed lines) and after (solid lines) fine-tuning. Given the same number of cluster centers, the feature dimension of \netfv representation is twice as large as \netvlad.}
\end{figure*}

\begin{figure*}[!htbp]
\begin{center}
\begin{tabular}{@{}ccc@{}}
\includegraphics[width=0.32\linewidth]{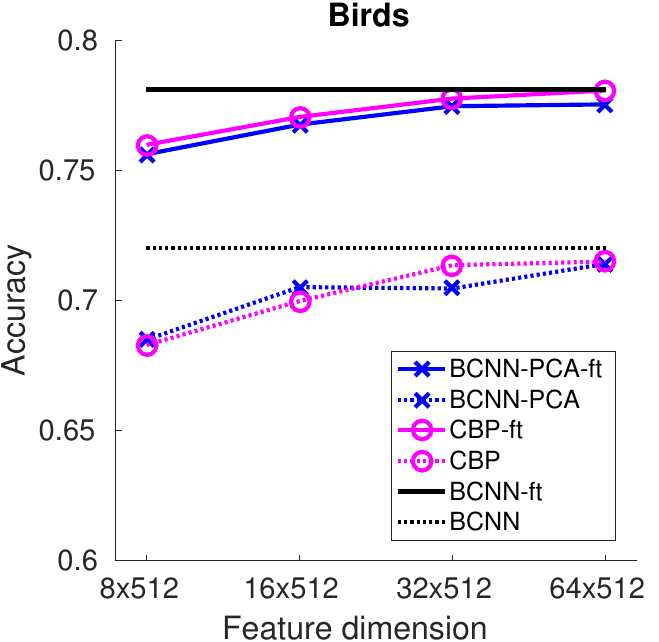} &
\includegraphics[width=0.32\linewidth]{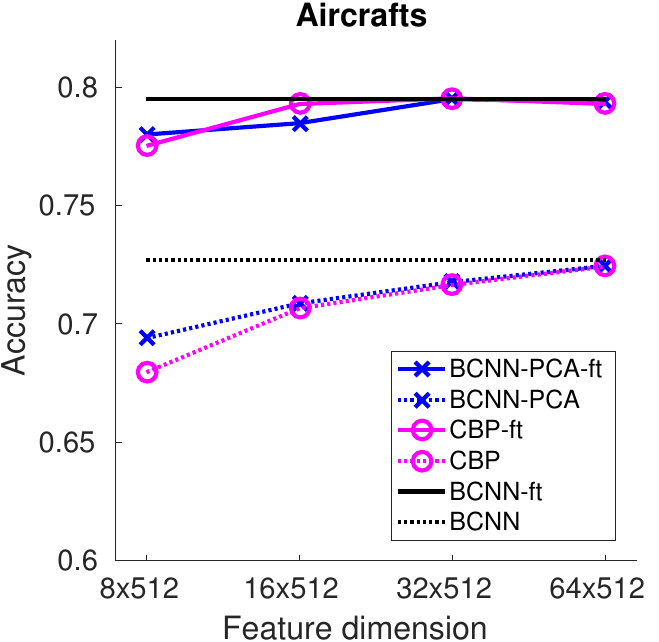} &
\includegraphics[width=0.32\linewidth]{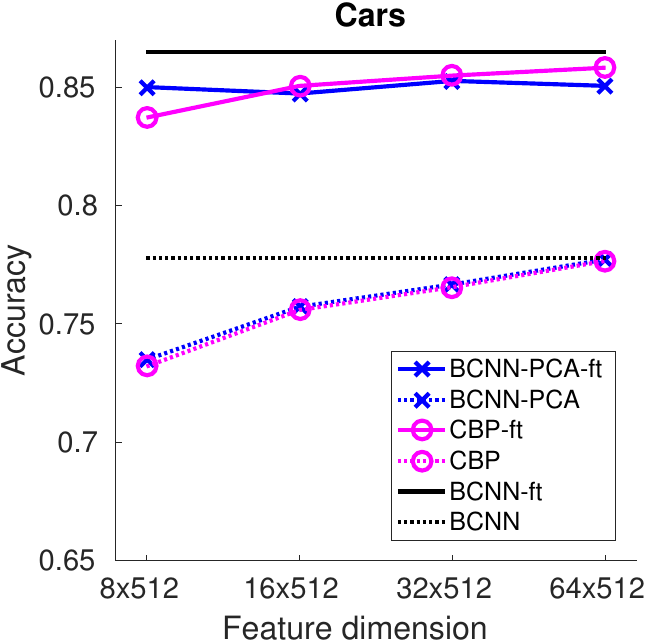} \\
\end{tabular}
\end{center}
\caption{\label{fig:bcnn_dim} Performance of B-CNNs using VGG-M \textit{relu5} features as function of feature dimension before (dashed lines) and after (solid lines) fine-tuning. One of the $512$ dimensional feature is projected using PCA to $k$ dimensions leading to a outer product of size $k \times 512$ (see Section~\ref{sec:dim_red} for details). The performance using Compact Bilinear Pooling (CBP)~\cite{Gao2016CVPR} and the full $512 \times 512$-dimensional model is shown in red and black respectively.}
\vspace{-0.2in}
\end{figure*}

In this section we investigate the trade-off between accuracy and feature dimension for various texture models proposed in Section~\ref{s:model} for fine-grained recognition. For \netvlad and \netfv the feature dimension can be varied by changing the number of cluster centers. For B-CNNs, consider the case where the outer product is computed among features $\mathbf{x}$ and $\mathbf{y}$. There are several strategies for reducing the feature dimension:
\begin{enumerate}[(1)]
\item Projecting the outer product into a lower dimensional space, \ie, $\Phi(\mathbf{x},\mathbf{y}) = \texttt{vec}(\mathbf{x}^T\mathbf{y}) \mathbf{P}$, where $\mathbf{P}$ is a projection matrix and the $\texttt{vec}$ operator reshapes the matrix into a vector.
 
\item Projecting both the features into a lower-dimensional space and computing outer product, $\Phi(\mathbf{x},\mathbf{y}) = (\mathbf{x}\mathbf{A}) ^T(\mathbf{y}\mathbf{B})$, where $\mathbf{A}$, $\mathbf{B}$ are projection matrices.
 
\item Projecting one of the features into a lower-dimensional space and computing the outer product, \ie, by setting $\mathbf{B}$ to an identity matrix in the previous approach.
\end{enumerate}

In each case, the projection matrices can be initialized using Principal Component Analysis (PCA). Although the first approach is straightforward, computing the PCA is computationally expensive due to the high dimensionality of the features (the covariance matrix of the outer product has $d^4$ entries for $d$-dimensional features). The second approach is computationally attractive but the outer product of two PCA projected features results in a  significant reduction in accuracy as shown in our earlier work~\cite{lin2015bilinear}, and more recently in~\cite{Gao2016CVPR}. We believe this is because after the PCA rotation, the features are no longer correlated across dimensions. 
Remarkably, reducing the dimension of only one feature  using PCA (third option) works well in practice.
While the projection can be initialized using PCA, they can be trained jointly with the classification layers. This technique was used in our earlier work~\cite{lin2015bilinear} to reduce the feature size. It breaks the symmetry of the features when identical networks are used and is an example of a partially shared feature pipeline (Figure~\ref{fig:two_streams}b). It also resembles the computations of VLAD and FV representations where both $f_A$ and $f_B$ are based on the same underlying feature.

The accuracy as a function of feature dimension shown in Figure~\ref{fig:gmm_components} for NetFV and NetVLAD. These results are obtained by varying the number of cluster centers. The results indicate the performance of NetVLAD and NetFV do not improve with more cluster centers beyond 32.

Figure~\ref{fig:bcnn_dim} shows the same for B-CNN features. We also compare our PCA approach to the recently-proposed Compact Bilinear Pooling (CBP) technique~\cite{Gao2016CVPR}. CBP approximates the outer product using a product of sparse linear projections of features with a Tensor Sketch~\cite{pham2013fast}. The performance of the full model with $512\times 512$ dimensions with and without fine-tuning is shown as a straight line. On birds and aircrafts the dimensionality can be reduced by 16$\times$ (\ie, to 32$\times$512) with a less than 1\% loss in accuracy. In comparison, NetVLAD with the same feature size (\ie, with 32 components) is about 3-4\% less accurate. Overall, for a given budget of dimensions the projected B-CNNs outperform \netvlad  and \netfv representations. The PCA approach is slightly worse than CBP. However, one advantage is that PCA can be implemented as a dense matrix multiplication which is empirically 1.5$\times$ faster than CBP which involves computing Fourier transforms and their inverses.

\subsection{Training B-CNNs on ImageNet LSVRC}
\label{s:imagenet}
We evaluate B-CNNs trained \emph{from scratch} on the ImageNet LSRVC 2012 dataset~\cite{russakovsky15imagenet}. In particular, we train a \bcnn with a \emph{relu5} layer output of VGG-M network and compare it to the standard VGG-M network. This allows a direct comparison of bilinear pooling with FC pooling since the rest of the architecture is identical in both networks. Additionally, we compare the effect of implicit translational invariance obtained in CNNs by spatially "jittering" the data, as well a explicit translation invariance of B-CNNs due to the orderless pooling. 

We train the two networks to classify 224$\times$224 images with different amounts of spatial jittering -- ``f1" for flip, ``f5" for flip + 5 translations, and ``f25" for flip + 25 translations. Training is done with stochastic sampling where one of the jittered copies is randomly selected for each example. The parameters are randomly initialized and trained using stochastic gradient descent with momentum for a number of epochs. We start with a high learning rate and reduce it by a factor of 10 when the validation error stops decreasing, and continue till no further improvement is observed.

Table~\ref{tab:imagenet} shows the ``top1" and ``top5" validation errors for \bcnn and VGG-M. The validation error is reported on a single center cropped image. Note that we train all networks with neither PCA color jittering nor batch normalization and our baseline results are within $2\%$ of the top1 errors reported in~\cite{chatfield14return}. The VGG-M model achieves $46.4\%$ top1 error with flip augmentation during training. The performance improves significantly to $39.6\%$ with f25 augmentation. \bcnn achieves $38.7\%$ top1 error with f1 augmentation, outperforming VGG-M trained with f25 augmentation. The results show that \bcnn  is discriminative, robust to translation and that explicit translation invariance is more effective. This trend is also reflected in the latest deep architectures such as Residual Networks~\cite{He_2016_CVPR} that replace the fully-connected layers with global pooling layers.

\begin{table}
\caption{\label{tab:imagenet} Accuracy on the ILSVRC 2014 \emph{validation} set using bilinear and FC pooling on top of \emph{relu5} layer output of a VGG-M network. Both networks are trained from scratch on the ILSVRC 2012 training set with varying amounts of data augmentation.}
\small
\renewcommand{\arraystretch}{1.5}
\setlength{\tabcolsep}{7pt}
\vspace{-0.2in}
\begin{center}
\begin{tabular}{l|ccc|cc}
 & \multicolumn{3}{c|}{\bf{Bilinear pooling}} & \multicolumn{2}{c}{\bf{FC pooling}}  \\ 
\hline 
\bf{data aug.} & f1 & f5 & f25 & f1 & f25 \\ 
\hline 
\it{error@1} & $38.7$  & $37.1$ & $36.6$  & $46.4$ & $39.6$ \\ 
\it{error@5} & $17.0$  & $16.3$  & $16.0$ & $22.5$ & $17.6$ \\ 
\end{tabular}
\end{center}
\vspace{-0.2in}
\end{table}

%\subsection{Augmenting spatial information}
%\label{sec:exp_spatial}
%\input{spatial}

%\subsection{Supervision vs. data vs. better models}
%
%\begin{enumerate}
%\item \todo{Result with bounding box? The trends when bounding-boxes are used at training and test times are similar. All the methods benefit from the added supervision. The performance of the FC and FV models improves significantly -- roughly 10\% for the FC and FV models with the M-Net and 6\% for those with the D-Net. However, the most accurate \bcnn model benefits less than 1\% suggesting a greater invariance to the location of parts in the image. 
%}
%\item \todo{Refer to the unreasonable effective of noisy data?}
%\item \todo{Result with resNet here?}
%\end{enumerate}

\subsection{Visualizing learned models}\label{s:viz}
\textbf{Top activations of B-CNN units.}
Figure~\ref{fig:bcnn-vis} shows the top activations of several units of the \emph{relu5\_3} layer of VGG-D and the \emph{relu5} layer of VGG-M network for the fine-tuned \bcnn (VGG-D + VGG-M) model. Both these networks contain units that activate strongly on highly localized features. For example, the last row of VGG-D detects ``tufted heads", while the fourth row in the same column detects a ``red-yellow stripe" on birds. Similarly for airplanes, the units localize different types of windows, noses, vertical stabilizers, with some specializing in detecting particular airliner logos. For cars, units activate on different kinds of head/tail lights, wheels, \etc

%The above visualizations also suggests that the role of features and parts in fine-grained recognition tasks can be traded. For instance, consider the task of gender recognition. One approach is to first train a gender-neutral face detector and followed by a gender classifier. However, it may be better to train a gender-specific face detector instead. By jointly training $f_A$ and $f_B$ the bilinear model can effectively trade-off the representation power of the features based on the data. Thus, manually defined parts not only requires significant annotation effort but also is likely to be sub-optimal when enough training data is available.

\begin{figure}
\begin{center}
\setlength{\tabcolsep}{2pt}
\renewcommand{\arraystretch}{0.5}
\begin{tabular}{cc}
    \multicolumn{2}{c}{Birds} \\
    \includegraphics[width=0.485\linewidth]{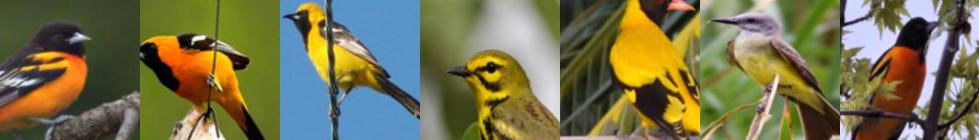} & \includegraphics[width=0.485\linewidth]{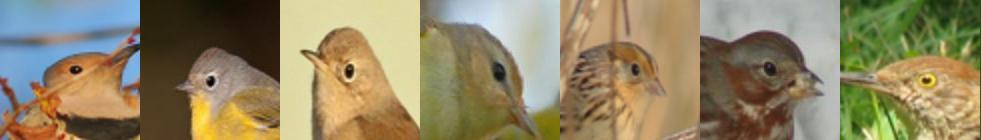} \\
    \includegraphics[width=0.485\linewidth]{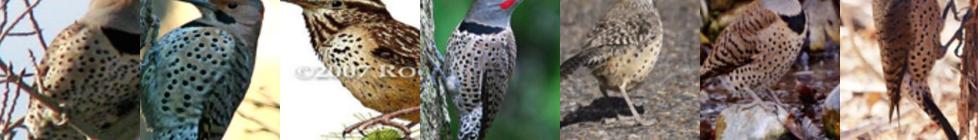} & \includegraphics[width=0.485\linewidth]{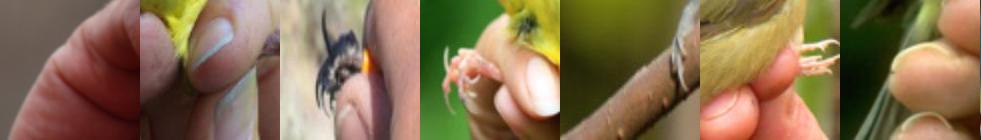} \\
    \includegraphics[width=0.485\linewidth]{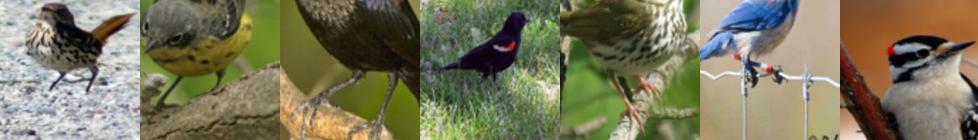} & \includegraphics[width=0.485\linewidth]{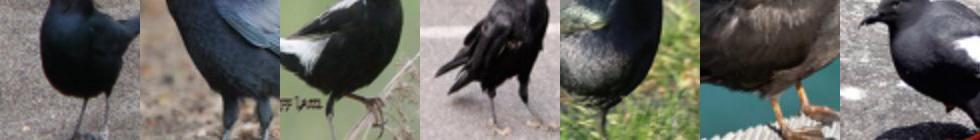} \\
    \includegraphics[width=0.485\linewidth]{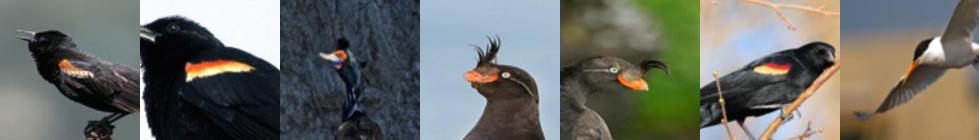} &  \includegraphics[width=0.485\linewidth]{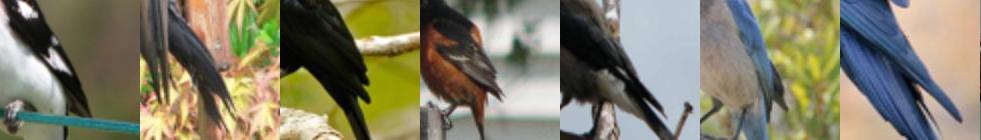}\\
    \includegraphics[width=0.485\linewidth]{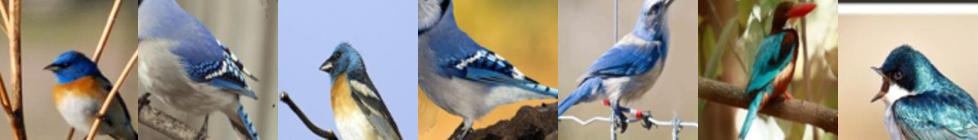} & \includegraphics[width=0.485\linewidth]{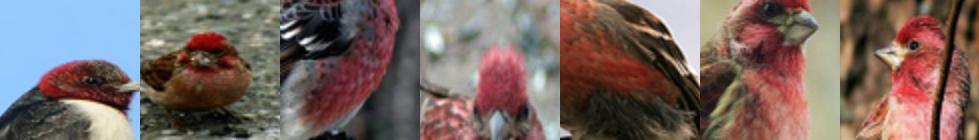}\\
    \includegraphics[width=0.485\linewidth]{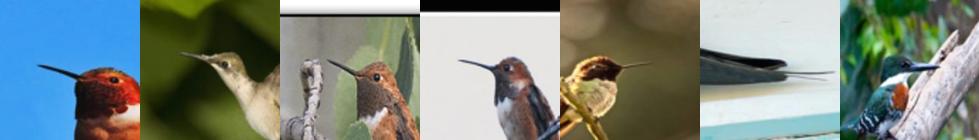} & \includegraphics[width=0.485\linewidth]{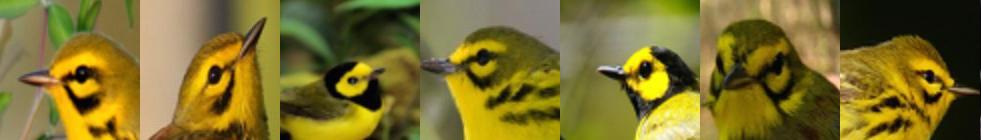}\\
    \includegraphics[width=0.485\linewidth]{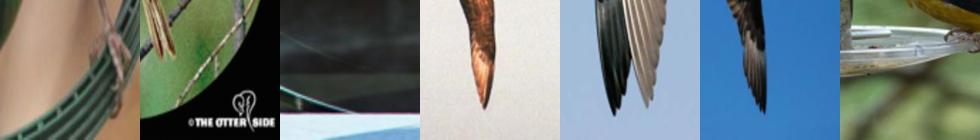} & \includegraphics[width=0.485\linewidth]{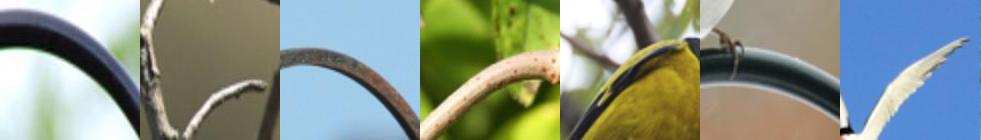}\\
    \includegraphics[width=0.485\linewidth]{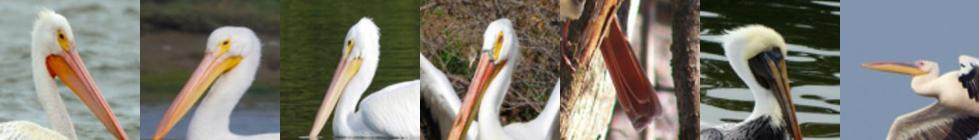} &  \includegraphics[width=0.485\linewidth]{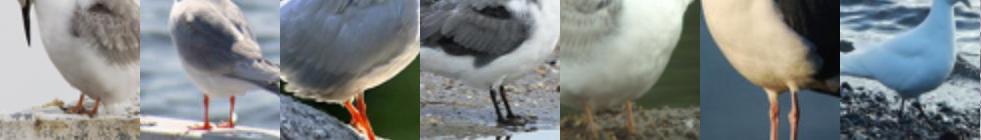}\\
    \includegraphics[width=0.485\linewidth]{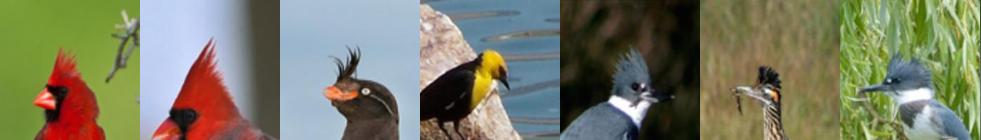} &  \includegraphics[width=0.485\linewidth]{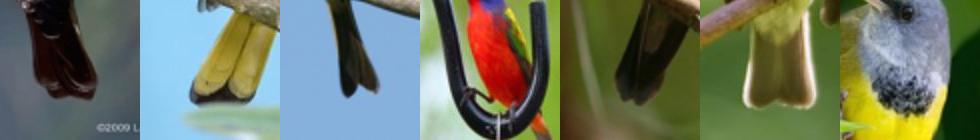}\\
    \includegraphics[width=0.485\linewidth]{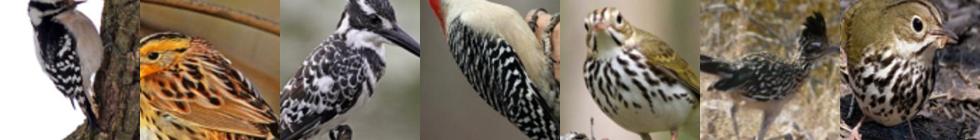} &  \includegraphics[width=0.485\linewidth]{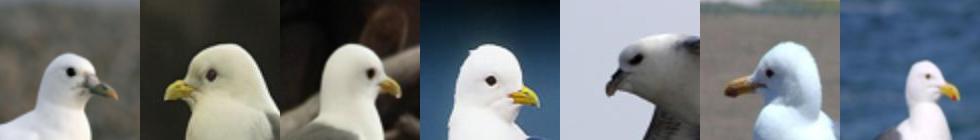}\\
    VGG-D filters & VGG-M filters \\
\end{tabular}
\begin{tabular}{cc}
\vspace{0.1cm} \\
    \multicolumn{2}{c}{Aircrafts} \\
    \includegraphics[width=0.485\linewidth]{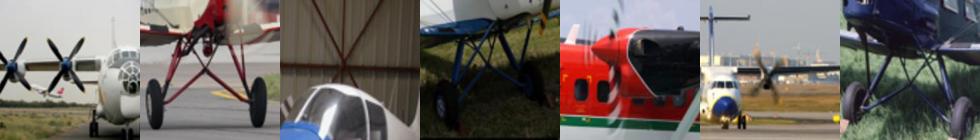} &   \includegraphics[width=0.485\linewidth]{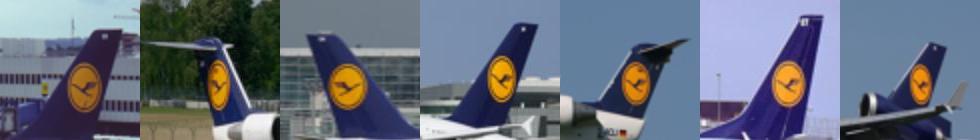} \\
    \includegraphics[width=0.485\linewidth]{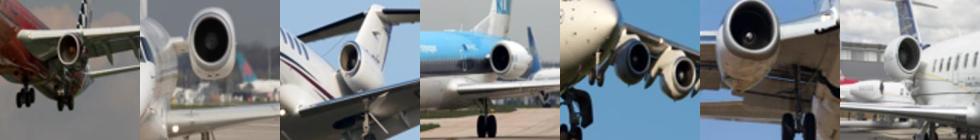} & \includegraphics[width=0.485\linewidth]{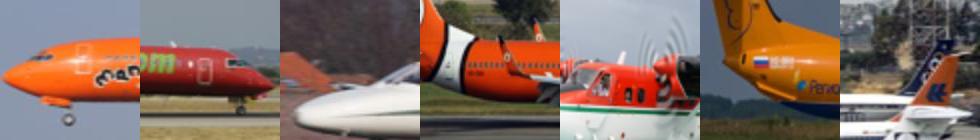} \\
    \includegraphics[width=0.485\linewidth]{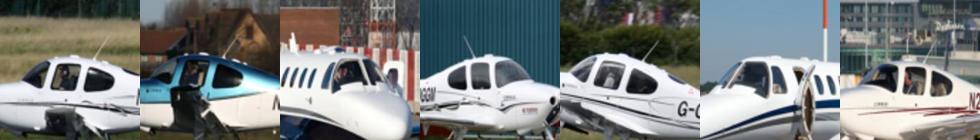} & \includegraphics[width=0.485\linewidth]{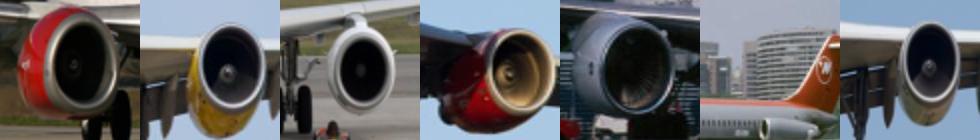} \\
    \includegraphics[width=0.485\linewidth]{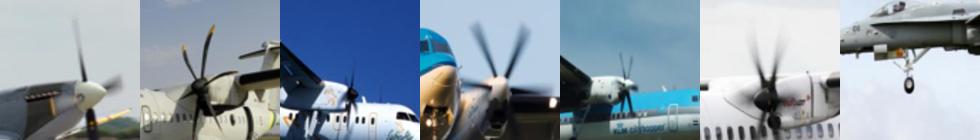} &  \includegraphics[width=0.485\linewidth]{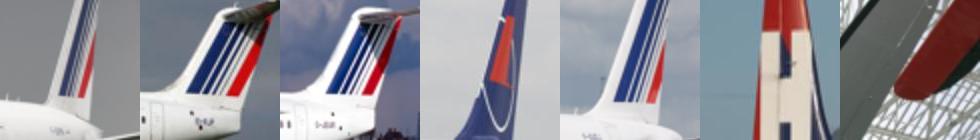}\\
    \includegraphics[width=0.485\linewidth]{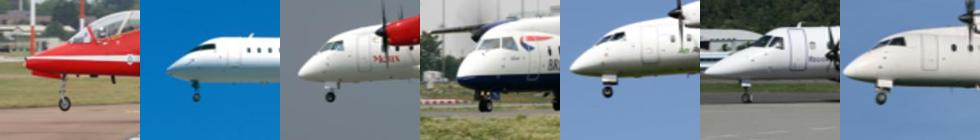} & \includegraphics[width=0.485\linewidth]{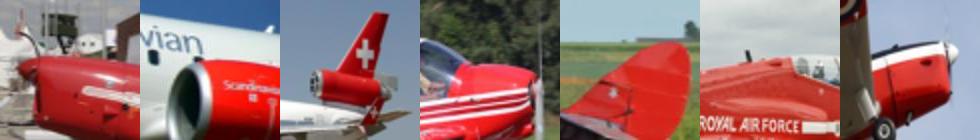}\\
    \includegraphics[width=0.485\linewidth]{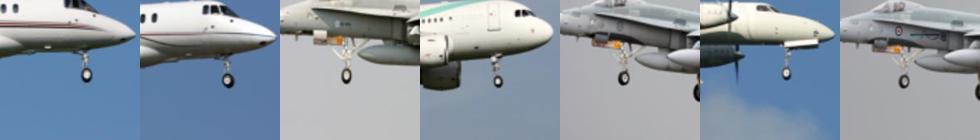} & \includegraphics[width=0.485\linewidth]{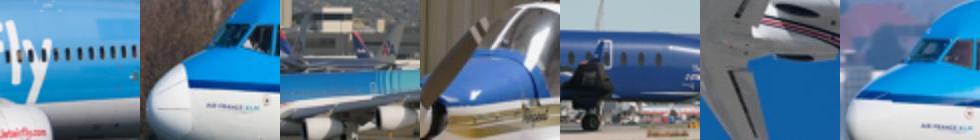}\\
    \includegraphics[width=0.485\linewidth]{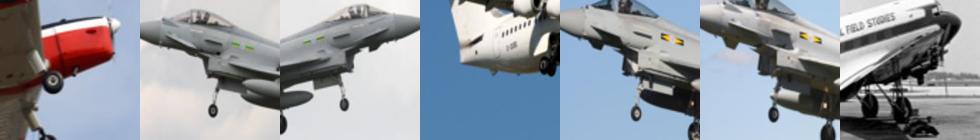} & \includegraphics[width=0.485\linewidth]{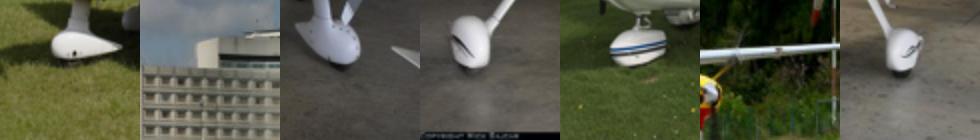}\\
    \includegraphics[width=0.485\linewidth]{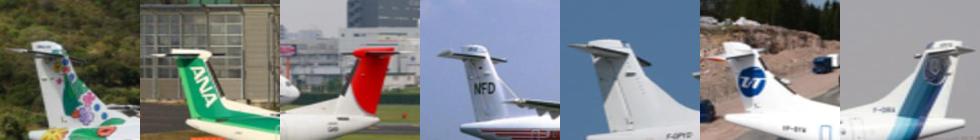} &  \includegraphics[width=0.485\linewidth]{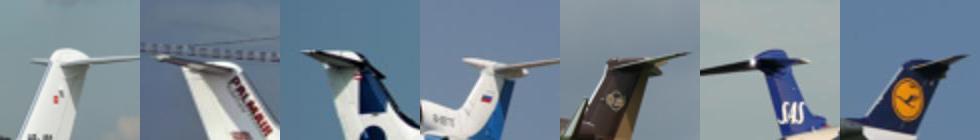}\\
    \includegraphics[width=0.485\linewidth]{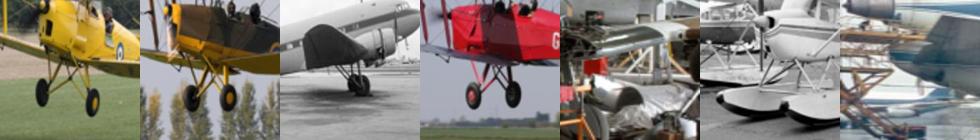} &  \includegraphics[width=0.485\linewidth]{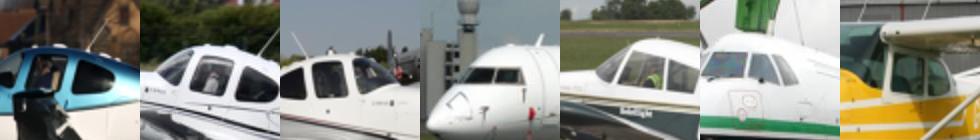}\\
    \includegraphics[width=0.485\linewidth]{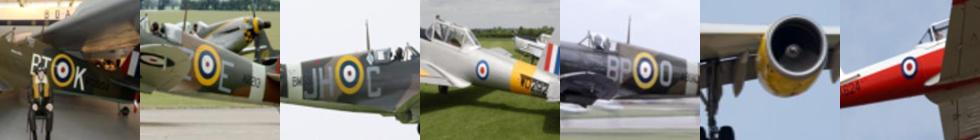} &  \includegraphics[width=0.485\linewidth]{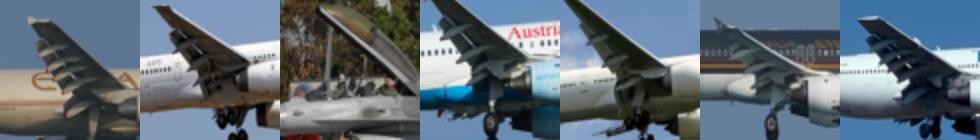}\\
    VGG-D filters & VGG-M filters \\
\end{tabular}
\begin{tabular}{cc}
\vspace{0.1cm} \\
    \multicolumn{2}{c}{Cars} \\
    \includegraphics[width=0.485\linewidth]{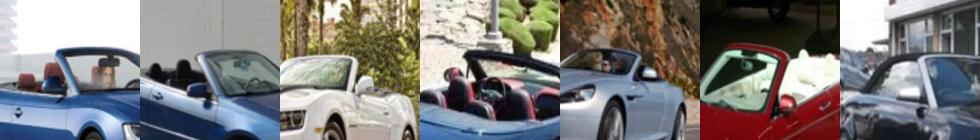} & \includegraphics[width=0.485\linewidth]{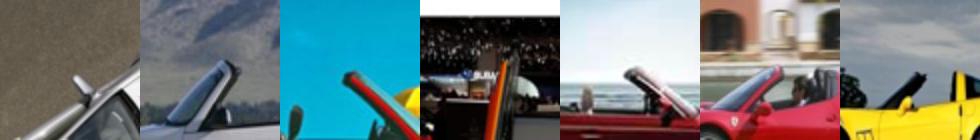} \\
    \includegraphics[width=0.485\linewidth]{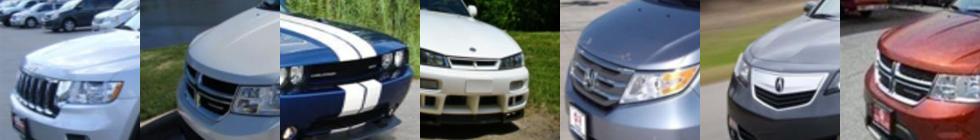} & \includegraphics[width=0.485\linewidth]{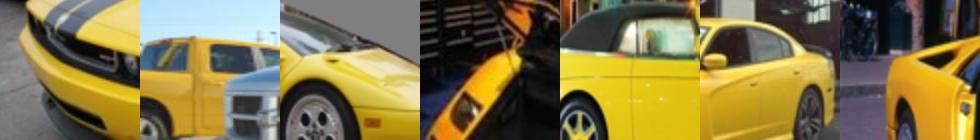} \\
    \includegraphics[width=0.485\linewidth]{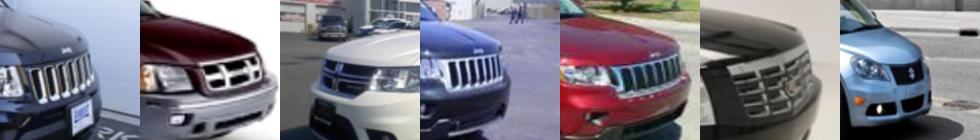} & \includegraphics[width=0.485\linewidth]{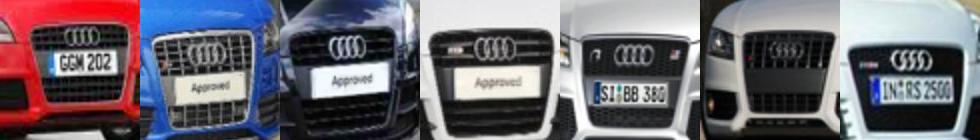} \\
    \includegraphics[width=0.485\linewidth]{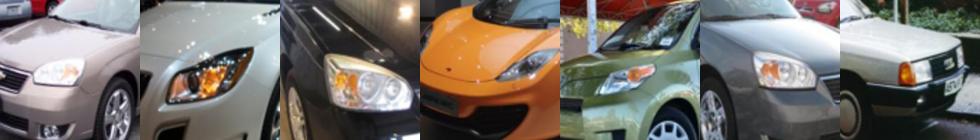} &  \includegraphics[width=0.485\linewidth]{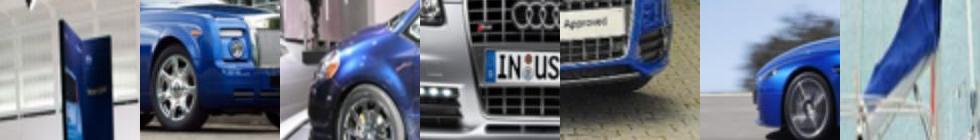}\\
    \includegraphics[width=0.485\linewidth]{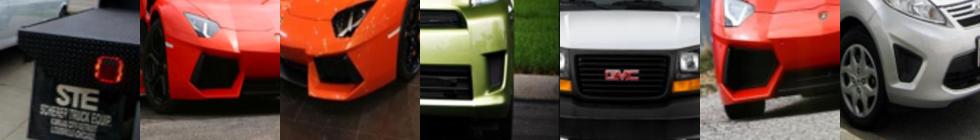} & \includegraphics[width=0.485\linewidth]{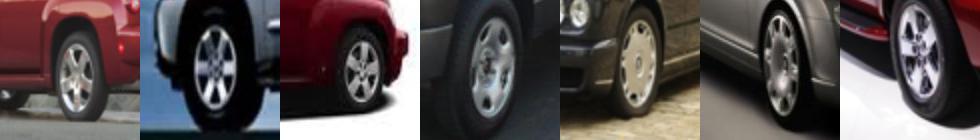}\\
    \includegraphics[width=0.485\linewidth]{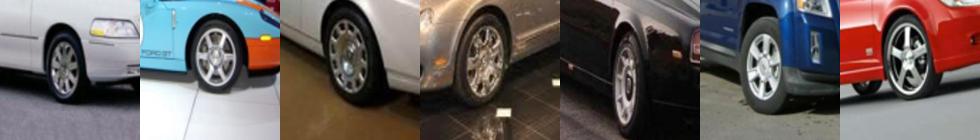} & \includegraphics[width=0.485\linewidth]{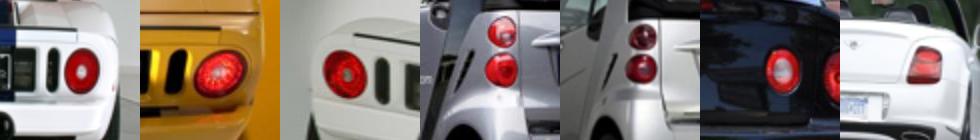}\\
    \includegraphics[width=0.485\linewidth]{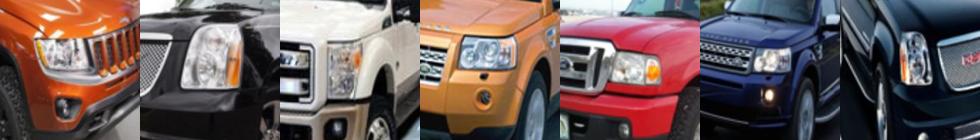} & \includegraphics[width=0.485\linewidth]{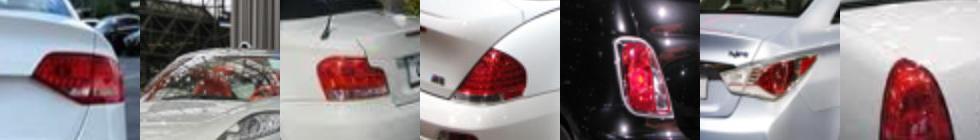}\\
    \includegraphics[width=0.485\linewidth]{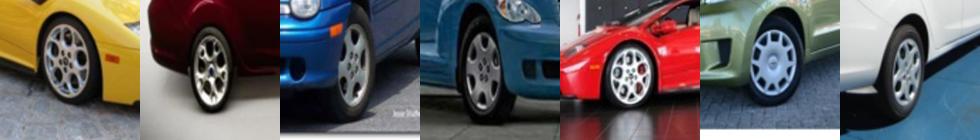} &  \includegraphics[width=0.485\linewidth]{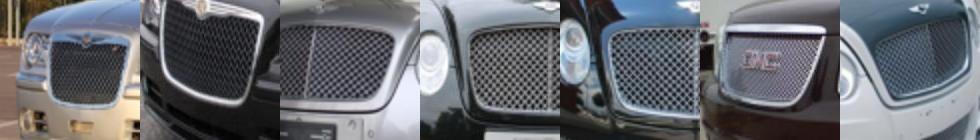}\\
    \includegraphics[width=0.485\linewidth]{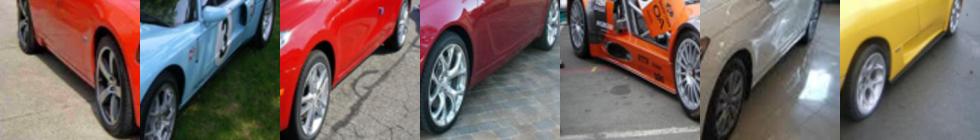} &  \includegraphics[width=0.485\linewidth]{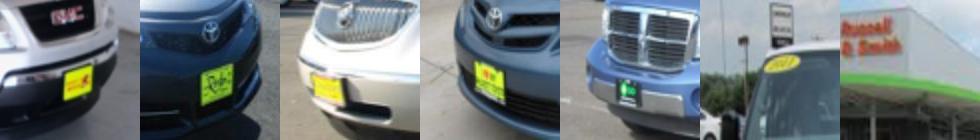}\\
    \includegraphics[width=0.485\linewidth]{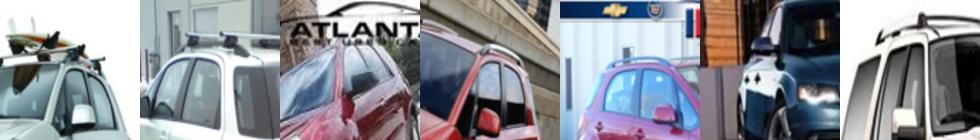} &  \includegraphics[width=0.485\linewidth]{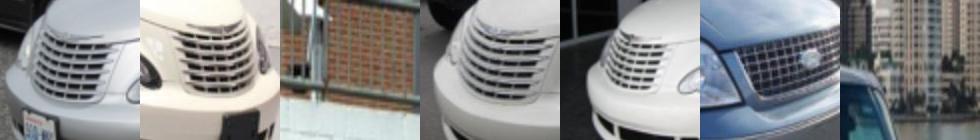}\\
    VGG-D filters & VGG-M filters \\
\end{tabular}

\caption{\label{fig:bcnn-vis} Patches with the highest activation for several filters of the fine-tuned \bcnn (VGG-D + VGG-M) model on birds, aircrafts and cars dataset. These visualizations indicate that network units activate on highly localized attributes of objects that capture color, texture, and shape patterns.}
\end{center}
\end{figure}

\begin{figure*}[th]
\begin{center}
\renewcommand{\arraystretch}{0.8}
\setlength{\tabcolsep}{1pt}
\begin{tabular}{cccccc}
%bcnn
\puti{braided}{\includegraphics[width=0.1615\linewidth]{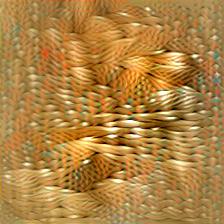}} & 
\puti{honeycombed}{\includegraphics[width=0.1615\linewidth]{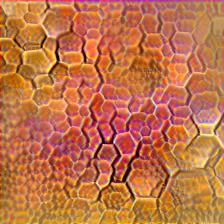}} & 
\puti{foliage}{\includegraphics[width=0.1615\linewidth]{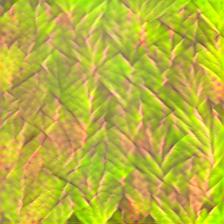}} &
\puti{water}{\includegraphics[width=0.1615\linewidth]{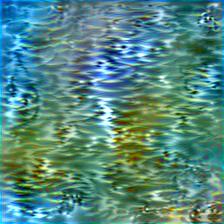}} &
\puti{bookstore}{\includegraphics[width=0.1615\linewidth]{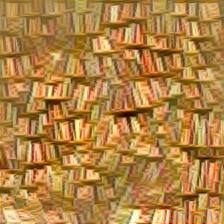}} &
\puti{bowling}{\includegraphics[width=0.1615\linewidth]{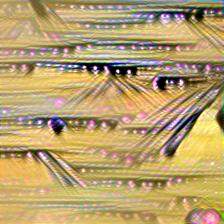}} \\ 
\puti{cobwebbed}{\includegraphics[width=0.1615\linewidth]{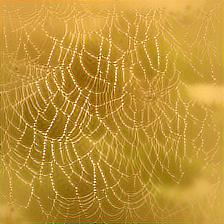}} & 
\puti{dotted}{\includegraphics[width=0.1615\linewidth]{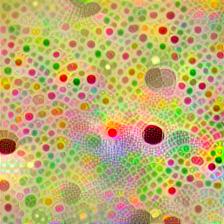}} & 
\puti{glass}{\includegraphics[width=0.1615\linewidth]{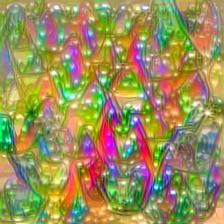}} &
\puti{wood}{\includegraphics[width=0.1615\linewidth]{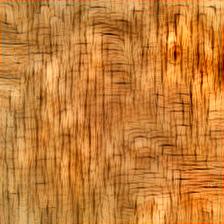}} &
\puti{classroom}{\includegraphics[width=0.1615\linewidth]{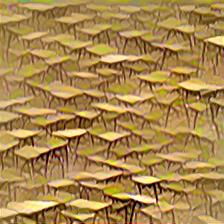}} &
\puti{closet}{\includegraphics[width=0.1615\linewidth]{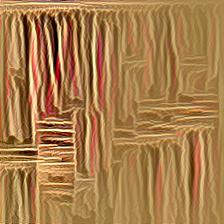}} \\ 
\puti{bumpy}{\includegraphics[width=0.1615\linewidth]{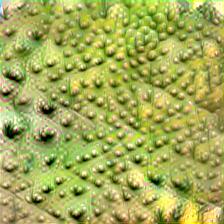}} & 
\puti{cracked}{\includegraphics[width=0.1615\linewidth]{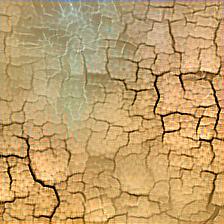}} & 
\puti{metal}{\includegraphics[width=0.1615\linewidth]{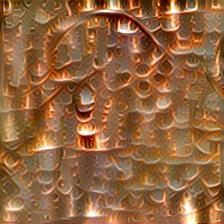}} &
\puti{paper}{\includegraphics[width=0.1615\linewidth]{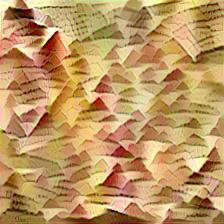}} &
\puti{laundromat}{\includegraphics[width=0.1615\linewidth]{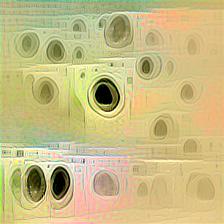}} &
\puti{florist}{\includegraphics[width=0.1615\linewidth]{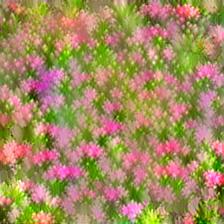}} \\ 
\puti{crested auklet}{\includegraphics[width=0.1615\linewidth]{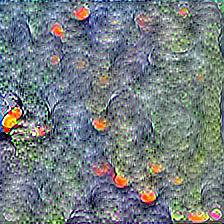}} & 
\puti{orchard oriole}{\includegraphics[width=0.1615\linewidth]{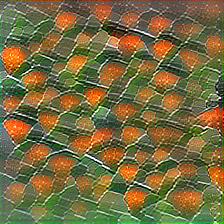}} & 
\puti{vermilion flycatcher}{\includegraphics[width=0.1615\linewidth]{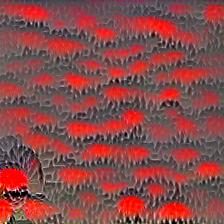}} &
\puti{redwinged blackbird}{\includegraphics[width=0.1615\linewidth]{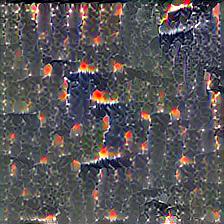}} &
\puti{northern flicker}{\includegraphics[width=0.1615\linewidth]{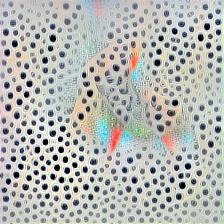}} &
\puti{chuck will widow}{\includegraphics[width=0.1615\linewidth]{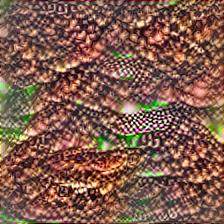}} \\ 
\puti{western grebe}{\includegraphics[width=0.1615\linewidth]{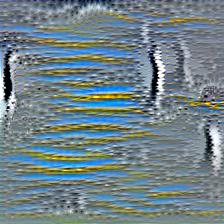}} & 
\puti{pine grosbeak}{\includegraphics[width=0.1615\linewidth]{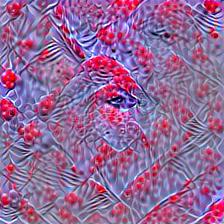}} & 
\puti{green violetear}{\includegraphics[width=0.1615\linewidth]{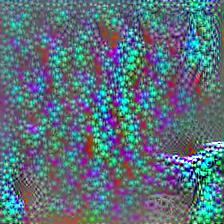}} &
\puti{baltimore oriole}{\includegraphics[width=0.1615\linewidth]{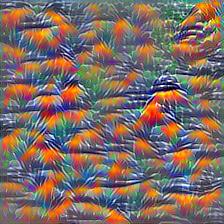}} &
\puti{canada warbler}{\includegraphics[width=0.1615\linewidth]{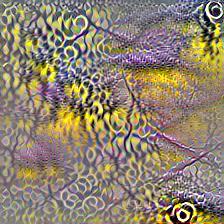}} &
\puti{downy woodpecker}{\includegraphics[width=0.1615\linewidth]{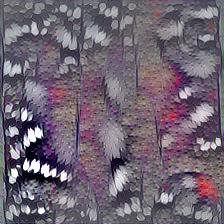}} \\ 
\end{tabular}
\end{center}
\vspace{-0.1in}
\caption{\label{fig:inverse} Visualizing various categories by inverting the B-CNN based on VGG-D network trained on DTD~\cite{cimpoi14describing}, FMD~\cite{sharan09material}, MIT Indoor dataset~\cite{quattoni09recognizing} (first three rows, two columns each from left to right), and the CUB dataset~\cite{WahCUB_200_2011} (last two rows, all columns). \emph{Best viewed in color and with zoom.}}
\vspace{-0.15in}
\end{figure*}

\vspace{0.15in}

\noindent
\textbf{Inverting categories.} To understand the properties learned by the B-CNNs we visualize pre-images of a category by ``inversion". We use the framework of Mahendran and Vedaldi~\cite{mahendran16visualizing} to generate an image that produces a high score for a target category $\hat{C}$ based on a B-CNN classifier. Specifically, for an image $\mathbf{x}$ and a layer $r_i,i=1,\ldots, n$, we compute the bilinear features $B_{r_i}$ using a B-CNN. Let $C_{r_i}$ be the class prediction probabilities obtained using linear classifier trained on $B_{r_i}$ in a supervised manner. We obtain an image that maximizes a target label $\hat{C}$ by solving the following optimization:

%one can compute the B-CNN features at a given layer $r_i$ to obtain a set of features $F_{r_i} = \{f_j\}$ indexed by their location $j$ and map the features $f_j$ to $g_A(f_j)$ and $g_B(f_j)$ (See Table~\ref{t:bilinear_approx} for the choice of $g_A$ and $g_B$). The bilinear feature $B_{r_i}({\cal I})$ of  $F_{r_i}$ is obtained by computing the outer product of features $g_A(f_j)$ and $g_B(f_j)$ and aggregating them across locations by averaging, \ie, 
%\begin{equation}
%B_{r_i}({\cal I}) = \frac{1}{N} \sum_{j=1}^N g_A(f_j)^T g_B(f_j)
%\end{equation}
%and normalizing the features (signed square-root and $\ell_2$). Let $r_i,i=1,\ldots, n,$ be the index of the $i^{th}$ layer with the bilinear feature representation $B_{r_i}$  from which we obtain category prediction probabilities $C_{r_i}$ by training a linear classifier in a supervised manner. Given a target category $\hat{C}$ we can obtain an image that matches the target label by solving the following optimization:

\begin{equation}
\label{eq:obj}
\min_\mathbf{x}\sum_{i=1}^{m} L\left( C_{r_i}, \hat{C}\right) + \gamma \Gamma(\mathbf{x}).
\end{equation}
Here, $L$ is a loss function such as the \emph{negative log-likelihood} of the label $\hat{C}$ and $\gamma$ is a tradeoff parameter. The image prior $\Gamma(\mathbf{x})$ encourages the smoothness of output image. We use the $\text{TV}_\beta$ norm with $\beta=2$: 
\begin{equation}
   \Gamma(\mathbf{x}) = \sum_{i,j} \left((x_{i,j+1} - x_{i,j})^2 + (x_{i+1,j} - x_{i,j})^2\right)^{\frac{\beta}{2}}.
\end{equation}
The exponent $\beta=2$ was empirically found to lead to fewer ``spike" artifacts in the optimization~\cite{mahendran16visualizing}. We use B-CNNs based on the VGG-D network. In our experiments, an input image is resized to 224$\times$224 pixels before computing the target bilinear features. We solve for $\mathbf{x} \in \mathbb{R}^{224\times 224\times 3}$ on the optimization. The lower resolution is primarily for speed since the dimension of the bilinear features are independent of the size of the image. We optimize the log-likelihood of the class probability using classifiers trained on bilinear features at \emph{relu2\_2, relu3\_3, relu4\_3, relu5\_3}. We use L-BFGS for optimization and compute the gradients of the objective with respect to $\mathbf{x}$ using back-propagation. The hyperparameter $\gamma=10^{-8}$ was found empirically to lead to good inverse images. We also refer readers to our earlier work~\cite{Lin2016CVPR} and more recent work~\cite{irmer17texture}, where this framework was applied for texture synthesis and style transfer using attributes.

%By choosing different mapping functions $g_A$ and $g_B$, we show the inverse images for various categories from various texture models. 

Figure~\ref{fig:inverse} shows some inverse images for various categories for the DTD, FMD, MIT indoor, and CUB-200-2011 dataset. These images reveal how B-CNNs represents various categories as textures. For instance, the \emph{dotted} category of DTD contains images of various colors and dot sizes and the inverse image is composed of multi-scale multi-colored dots. The inverse images of \emph{water} and \emph{wood} from FMD are highly representative of these categories. The inverse images of the MIT indoor dataset reveal key properties of a category -- a \emph{bookstore} has a racks of books, a \emph{laundromat} has laundry machines at various scales and locations. The inverse images of various bird species capture distinctive colors and patterns on their bodies. Figure~\ref{fig:texture-layers} visualizes reconstructions by incrementally adding layers in the bilinear representation. Even though the \emph{relu5\_3} layer provides the best recognition accuracy, simply using that layer did not produce good inverse images as the color information was missing.
%Combining all the layers leads to more visually interpretable inverses.

%In addition to \bcnn, Fig.~\ref{fig:inverse-comp} also shows examples of texture inverse obtained from \netbovw, \netvlad, and \netfv. We observe that the optimization is prone to bad local minima and produces less sharper outputs on these approximation models.

\begin{figure}
\begin{center}
\setlength{\tabcolsep}{1.5pt}
\renewcommand{\arraystretch}{0.7}
\begin{tabular}{cccc}
\textit{relu2\_2} & + \textit{relu3\_3} & + \textit{relu4\_3} & + \textit{relu5\_3} \\
\includegraphics[width=0.24\linewidth]{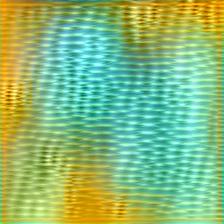} & 
\includegraphics[width=0.24\linewidth]{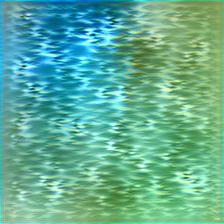} & 
\includegraphics[width=0.24\linewidth]{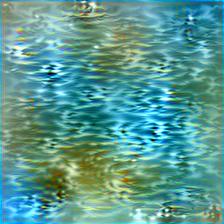} &
\puti{water}{\includegraphics[width=0.24\linewidth]{fig/viz/inv/fmd/water.jpg}} \\
\includegraphics[width=0.24\linewidth]{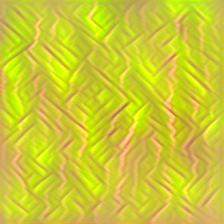} & 
\includegraphics[width=0.24\linewidth]{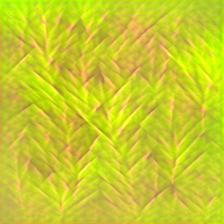} & 
\includegraphics[width=0.24\linewidth]{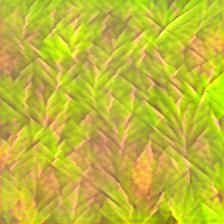} &
\puti{foliage}{\includegraphics[width=0.24\linewidth]{fig/viz/inv/fmd/foliage.jpg}} \\
\includegraphics[width=0.24\linewidth]{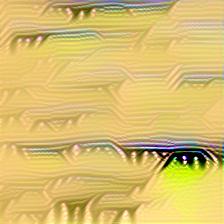} & 
\includegraphics[width=0.24\linewidth]{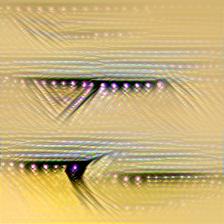} & 
\includegraphics[width=0.24\linewidth]{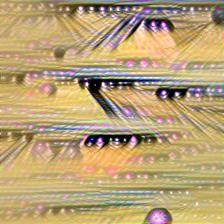} &
\puti{bowling}{\includegraphics[width=0.24\linewidth]{fig/viz/inv/mit_indoor/bowling.jpg}} \\
\end{tabular}
\end{center}
\vspace{-0.1in}
\caption{\label{fig:texture-layers} Inverse images obtained from a multilayer B-CNN. From left to right different layers (shown on top) are added one by one.}
\vspace{-0.2in}
\end{figure}

%\begin{figure}[th]
%\begin{center}
%\renewcommand{\arraystretch}{0.8}
%\setlength{\tabcolsep}{1pt}
%\begin{tabular}{cccccc}
%%bcnn
%\includegraphics[width=0.32\linewidth]{fig/viz/inv/dtd/honeycombed.png} & 
%\includegraphics[width=0.32\linewidth]{fig/viz/inv/fmd/foliage.png} &
%\puti{B-CNN}{\includegraphics[width=0.32\linewidth]{fig/viz/inv/mit_indoor/buffet.png}} \\ 
%%netvlad
%\includegraphics[width=0.32\linewidth]{fig/viz/invert-svm/dtd/netbovw/honeycombed.png} & 
%\includegraphics[width=0.32\linewidth]{fig/viz/invert-svm/fmd/netbovw/foliage.png} &
%\puti{NetBoVW}{\includegraphics[width=0.32\linewidth]{fig/viz/invert-svm/mit_indoor/netbovw/buffet.png}} \\ 
%%netfv 
%\includegraphics[width=0.32\linewidth]{fig/viz/invert-svm/dtd/netvlad/honeycombed.png} & 
%\includegraphics[width=0.32\linewidth]{fig/viz/invert-svm/fmd/netvlad/foliage.png} & 
%\puti{NetVLAD}{\includegraphics[width=0.32\linewidth]{fig/viz/invert-svm/mit_indoor/netvlad/buffet.png}} \\ 
%%netbovw 
%\includegraphics[width=0.32\linewidth]{fig/viz/invert-svm/dtd/netfv/honeycombed.png} & 
%\includegraphics[width=0.32\linewidth]{fig/viz/invert-svm/fmd/netfv/foliage.png} &
%\puti{NetFV}{\includegraphics[width=0.32\linewidth]{fig/viz/invert-svm/mit_indoor/netfv/buffet.png}} \\ 
%honeycombed & foliage & buffet \\
%\end{tabular}
%\end{center}
%\vspace{-0.1in}
%\caption{\label{fig:inverse-comp} Texture category inverses based on B-CNN, NetBoVW, NetVLAD, and NetFV models (from top to buttom). We refer readers to the page \url{http://maxwell.cs.umass.edu/viz} for the complete set of visualizations.}
%\end{figure}

\section{Conclusion}\label{s:conclusion}
We presented the B-CNN architecture that aggregates second-order statistics of CNN activations resulting in an orderless representation of an image. These networks can be trained in an end-to-end manner allowing both training from scratch on large datasets, and domain-specific fine-tuning for transfer learning. Moreover, these models are fairly efficient, processing 448$\times$448 resolution images at 30-100 FPS on a NVIDIA Titan X GPU. We also compared B-CNNs to both exact and approximate variants of deep texture representations and studied the accuracy and memory trade-offs they offer. The main conclusion was that variants of outer-product representations are highly effective at various fine-trained, texture, and scene recognition tasks. Moreover, these representations are redundant and in most cases their dimension can be reduced by an order of magnitude without significant loss in accuracy. A visualization of B-CNNs showed that these models effectively represent objects as \emph{texture} and their units are correlated with localized attributes useful for fine-grained recognition.

\vspace{0.1in}
\small{
\textbf{Acknowledgement:} This research was supported in part by the National Science Foundation grant IIS-1617917, a faculty gift from Facebook, and
by the Office of the Director of National Intelligence
(ODNI), Intelligence Advanced Research Projects Activity
(IARPA) under contract number 2014-14071600010. The
views and conclusions contained herein are those of the
authors and should not be interpreted as necessarily representing
the official policies or endorsements, either expressed
or implied, of ODNI, IARPA, or the U.S. Government.
The U.S. Government is authorized to reproduce and
distribute reprints for Governmental purpose notwithstanding
any copyright annotation thereon.
The experiments were performed using high performance 
computing equipment obtained under a grant from the Collaborative R\&D Fund 
managed by the Massachusetts Tech Collaborative and GPUs donated by NVIDIA.}

\bibliographystyle{ieee}
\bibliography{bibliography}
\vspace{-0.7in}
\begin{IEEEbiography}[{\includegraphics[width=1in,height=1.25in,clip,keepaspectratio]{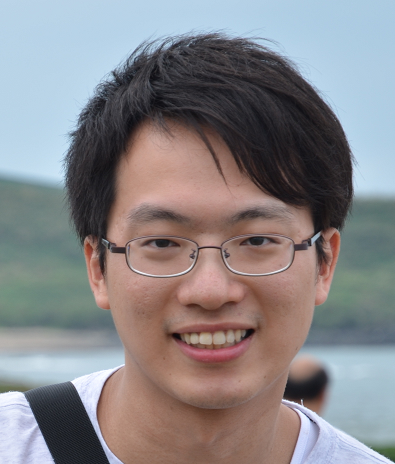}}]{Tsung-Yu Lin} received his BS and MS degrees in computer science from National Tsing Hua University in 2008 and 2010 respectively. He worked at Academia Sinica, a governmental research institute in Taiwan as research assistant after he obtained MS degree. He is currently a PhD student in the College of Information and Computer Sciences at University of Massachusetts Amherst. He is interesting in applying machine learning techniques to computer vision problems, especially image classification and segmentation.
\end{IEEEbiography}

\vspace{-0.7in}

\begin{IEEEbiography}[{\includegraphics[width=1in,height=1.25in,clip,keepaspectratio]{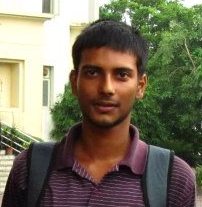}}]{Aruni RoyChowdhury} is a PhD student in the College of Information and Computer Sciences, University of Massachusetts, Amherst. His research interests are in Computer Vision, focusing on applications of deep learning to face recognition and fine-grained recognition. Previously he has worked on scene text detection and handwriting recognition at the Indian Statistical Institute, Kolkata. He obtained his MS degree (2016) from the University of Massachusetts, Amherst and his BTech (2013) from Heritage Institute of Technology, Kolkata in India.
\end{IEEEbiography}

\vspace{-0.6in}

\begin{IEEEbiography}[{\includegraphics[width=1in,height=1.25in,clip,keepaspectratio]{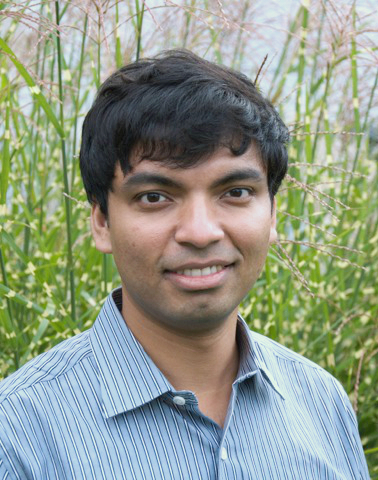}}]{Subhransu Maji} is an Assistant Professor in the College of Information and Computer Sciences at the University of Massachusetts, Amherst. Previously, he was a Research Assistant Professor at TTI Chicago, a philanthropically endowed academic computer science institute in the University of Chicago campus. He obtained his Ph.D. from the University of California, Berkeley in 2011, and B.Tech. in Computer Science and Engineering from IIT Kanpur in 2006. His research focuses on developing visual recognition architectures with an eye for efficiency and accuracy.
\end{IEEEbiography}

\end{document}